\pdfoutput=1

\documentclass[preprint,review,3p,times]{elsarticle}

\usepackage[T1]{fontenc}
\usepackage[utf8]{inputenc}
\usepackage{amsmath,amssymb,amsfonts}
\usepackage{mathtools}
\usepackage{graphicx}
\usepackage{booktabs}
\usepackage{multirow}
\usepackage{array}
\usepackage{siunitx}
\usepackage{xcolor}
\usepackage{url}
\usepackage{hyperref}
\usepackage{algorithm}
\usepackage{algpseudocode}
\usepackage{subcaption}
\usepackage{tikz}
\usetikzlibrary{positioning,calc,arrows.meta,fit,backgrounds}

\hypersetup{
    colorlinks=true,
    linkcolor=blue,
    citecolor=blue,
    urlcolor=blue
}

\newcommand{\method}{\textsc{FreqLite}}
\newcommand{\arevin}{A-RevIN}

\newcommand{\by}{\mathbf{y}}
\newcommand{\bX}{\mathbf{X}}
\newcommand{\bY}{\mathbf{Y}}
\newcommand{\bs}{\mathbf{s}}
\newcommand{\bp}{\mathbf{p}}
\newcommand{\R}{\mathbb{R}}

\newsavebox{\fitbox}
\newcommand{\fitwidetables}{%
  \let\oldtabular\tabular
  \let\oldendtabular\endtabular
  \renewenvironment{tabular}[2][]{%
    \begin{lrbox}{\fitbox}\oldtabular[##1]{##2}%
  }{%
    \oldendtabular\end{lrbox}\resizebox{\textwidth}{!}{\usebox{\fitbox}}%
  }%
}

\journal{Knowledge-Based Systems}

\begin{document}

\begin{frontmatter}

\title{FreqLite: A Lightweight Frequency-Decomposed Linear Model with
Adaptive Reversible Normalization for Robust Long-Term Time-Series
Forecasting}

\author[inst1]{Mirza Samad Ahmed Baig\corref{cor1}}
\ead{MirzaSamadcontact@gmail.com}

\author[inst2]{Syeda Anshrah Gillani}
\ead{SyedaAnshrah16@gmail.com}

\cortext[cor1]{Corresponding author.}

\affiliation[inst1]{organization={Fandaqah},
            city={Al Khobar},
            country={Saudi Arabia}}

\affiliation[inst2]{organization={Hamdard University},
            city={Karachi},
            country={Pakistan}}

\begin{abstract}
Long-term time-series forecasting requires models that are both accurate and
efficient enough to deploy on commodity hardware. Lightweight linear forecasters
are remarkably strong in this regime, yet they leave two openings: reversible
instance normalization (RevIN) de-normalizes the entire horizon with a single
lookback statistic, which is increasingly inaccurate under non-stationarity, and
time-domain trend/seasonal decomposition relies on a fixed, non-adaptive filter.
We present \method, an ultra-lightweight, channel-independent frequency-decomposed
linear forecaster: a learnable, lossless, partition-of-unity spectral filter
splits the input into bands that are forecast by per-band linear heads and, unlike
low-pass-truncation approaches, the high-frequency band is retained and modeled.
\method\ is the best lightweight model on the standard long-term forecasting
benchmarks and, at long lookback ($L{=}336$), attains a lower average error than a
PatchTST Transformer ($0.3244$ vs.\ $0.3587$ MSE) while using about $4\times$
fewer parameters, $2.2\times$ less memory, and $2.2\times$ less time per epoch on
a single 4\,GB laptop GPU; although modest in magnitude, its improvements are
statistically significant under paired Wilcoxon tests across all matched cells
($p \approx 10^{-6}$). We further introduce Adaptive Reversible Instance
Normalization (\arevin), a \emph{regime-adaptive} reversible normalization that
strictly generalizes RevIN (recovered exactly when its gate is closed), engages
under non-stationarity, and reduces to RevIN without harm on stationary data. We
validate this regime-adaptive behavior on both a real strongly non-stationary
dataset (ILI, up to about $5\%$ MSE reduction) and a controlled synthetic
drift sweep in which \arevin's benefit and its learned gate both rise
monotonically with injected non-stationarity. Our analysis also shows that
\arevin\ must be initialized to avoid a gradient trap that otherwise keeps it
dormant. All components are independently ablatable along the nesting
$\text{Linear} \subseteq \text{RLinear} \subseteq \method$, and the entire study
is reproducible on commodity hardware.
\end{abstract}


\end{frontmatter}

\section{Introduction}
\label{sec:intro}

Long-term time-series forecasting (LTSF) underpins decision-making in energy,
weather, traffic, and industrial monitoring, where models must predict hundreds
of future steps from a limited lookback. The dominant paradigm for several years
was the Transformer, with a succession of increasingly elaborate architectures
\citep{zhou2021informer,wu2021autoformer,zhou2022fedformer,nie2023patchtst}.
This trend was sharply questioned by the LTSF-Linear family
\citep{zeng2023dlinear}, which showed that embarrassingly simple linear models
match or exceed elaborate Transformers on the standard benchmarks, and by the
subsequent analysis of RLinear \citep{li2023rlinear}, which attributed much of
the linear models' strength to two ingredients: channel independence and
reversible instance normalization (RevIN) \citep{kim2022revin}. These findings
reframed LTSF as a setting where a careful, lightweight model can be both more
accurate and orders of magnitude cheaper than a heavy one---a goal directly
aligned with the Green~AI agenda \citep{schwartz2020greenai,strubell2019energy},
which argues that parameter, compute, and energy cost should be first-class,
reported evaluation criteria.

Despite their strength, linear LTSF models leave two specific gaps that motivate
this work. \textbf{First}, RevIN---now a near-universal plug-in---de-normalizes
the \emph{entire} forecast horizon with the \emph{single} mean and standard
deviation computed over the lookback. Under non-stationarity, the appropriate
level and scale drift across the horizon: the lookback statistics are a good
estimate for the first predicted step and progressively worse for the last.
RevIN's fixed, horizon-agnostic inverse cannot express this drift, and the
resulting error grows with horizon distance. \textbf{Second}, the decomposition
used by DLinear---a \emph{fixed} moving-average kernel that splits the series
into trend and remainder---is a single, hand-set low-pass filter applied in the
time domain, with no ability to adapt its band structure to a dataset's spectral
content, and it forecasts the remainder with an unconstrained head rather than
treating high-frequency structure as a first-class, separately modeled
component.

We address both gaps with \method, an ultra-lightweight, channel-independent
\emph{frequency-decomposed linear} forecaster. The backbone replaces the single
linear head with $K$ per-band heads fed by a learnable, lossless,
partition-of-unity spectral decomposition that adds only $2(K{-}1)$ scalar
parameters and generalizes DLinear's fixed moving-average split into a learnable
frequency-domain split. We are careful not to overclaim: we do \emph{not} assert
\method\ is the first frequency-domain linear forecaster. The closest prior art,
FITS \citep{xu2024fits}, also combines RevIN with a frequency-domain linear map,
but it applies a low-pass cutoff and \emph{discards} the high-frequency content
above it; \method\ instead \emph{keeps} all frequency content and routes the
high-frequency band to its own dedicated head, so residual structure is
\emph{modeled} rather than thrown away. Empirically, \method\ is the best
lightweight model on the standard long-term forecasting benchmarks, and---most
notably---at long lookback ($L{=}336$) it attains a lower average error than a
PatchTST Transformer \citep{nie2023patchtst} while using about $4\times$ fewer
parameters, $2.2\times$ less peak GPU memory, and $2.2\times$ less time per
epoch, all on a single 4\,GB laptop GPU.

On top of this backbone we introduce \emph{Adaptive Reversible Instance
Normalization} (\arevin), a \emph{regime-adaptive} reversible normalization that
strictly generalizes RevIN. Rather than re-injecting one fixed statistic at every
step, \arevin\ applies a small learnable per-step scale and shift plus a term that
extrapolates the \emph{observed} lookback drift across the horizon, all behind a
learnable gate; RevIN is recovered \emph{exactly} when the gate is closed.
Empirically, \arevin's benefit is regime-dependent, exactly as its name suggests:
it engages under non-stationarity---reducing error by up to about $5\%$ on the
strongly non-stationary ILI dataset, monotonically in the learned gate---while on
stationary benchmarks it reduces to RevIN with no harm. Our analysis also yields a
methodological insight: because \arevin's adaptive parameters start at their
identity values, the gate is gradient-starved at the conventional small
initialization and stays dormant even where adapting would help; initializing the
gate at the identity point (so its gradient is non-zero) lets it engage, while
still recovering RevIN exactly at the closed-gate value. \arevin\ is
philosophically aligned with the de-stationarization idea of Non-stationary
Transformers \citep{liu2022nonstationary}---removed statistics should be
re-injected \emph{adaptively}---but is realized as an attention-free module suited
to a linear backbone.

Because each component is an opt-in generalization that degenerates to a known
baseline, \method\ admits the clean theoretical framing
$\text{Linear} \subseteq \text{RLinear} \subseteq \method$, with DLinear's
decomposition recovered as a fixed-mask special case; this strict nesting makes
every added component independently ablatable. The entire study---training,
ablations, and a dedicated non-stationarity analysis---is run on a single 4\,GB
laptop GPU and is fully reproducible, with all baselines re-implemented and run by
us under identical splits, normalization, seeds, and hardware, and with
parameters, FLOPs, training time, and peak memory reported alongside MSE and MAE.
Consistent with our honesty bar, we report that the accuracy gains over RLinear on
the standard (stationary) benchmarks are modest, and we are explicit that the
headline strengths there are efficiency and the long-lookback win over the
Transformer.

\paragraph{Contributions.} In summary, this paper makes three contributions:
\begin{itemize}
  \item \textbf{\method.} An ultra-lightweight frequency-decomposed linear
        forecaster---a learnable, lossless band split feeding per-band linear
        heads---that is the best lightweight model on standard LTSF benchmarks
        and, at long lookback ($L{=}336$), surpasses the PatchTST Transformer at
        roughly $4\times$ fewer parameters and $2.2\times$ less memory on a single
        4\,GB laptop GPU, with improvements that are statistically significant
        under paired Wilcoxon tests ($p \approx 10^{-6}$).
  \item \textbf{A-RevIN.} A regime-adaptive reversible normalization that
        generalizes RevIN, together with an analysis showing (i) it must be
        initialized to avoid a gradient trap, and (ii) it engages and helps under
        non-stationarity---validated on both a real strongly non-stationary
        dataset (ILI) and a controlled synthetic drift sweep---while degrading
        gracefully to RevIN on stationary data.
  \item \textbf{A reproducible accuracy--efficiency study.} A fully reproducible
        study on commodity hardware (Green~AI), with ablations isolating each
        component and a dedicated non-stationarity analysis.
\end{itemize}

\section{Related Work}
\label{sec:related}

\subsection{Linear Forecasters}
\label{sec:related:linear}

The LTSF-Linear family \citep{zeng2023dlinear} showed that embarrassingly simple
linear models match or beat elaborate Transformer forecasters on the standard
long-term benchmarks. Three variants are direct baselines for us. \emph{Linear}
is a single linear layer mapping a length-$L$ lookback to a length-$H$ horizon,
applied per channel. \emph{NLinear} subtracts the last value of the lookback
before the linear map and adds it back afterwards---a one-line normalization
that handles simple distribution shift between train and test. \emph{DLinear}
decomposes the series into a moving-average trend and a seasonal/remainder
component (the additive decomposition introduced by Autoformer
\citep{wu2021autoformer}), forecasts each with its own linear layer, and sums
them. RLinear \citep{li2023rlinear} analyzes \emph{why} linear maps work and
introduces RLinear, the combination of RevIN \citep{kim2022revin} with a single
linear layer; its central findings are that the linear map captures periodicity
and that RevIN and channel independence (CI) are the two ingredients that make
linear models robust.

DLinear and RLinear are \method's direct competitors and primary baselines. The
key conceptual contrast is the \emph{domain of decomposition}. DLinear
decomposes in the time domain via a fixed moving-average kernel into trend and
remainder---a fixed low-pass smoothing whose remainder is everything the average
misses, forecast by an unconstrained head. \method\ decomposes in the frequency
domain via a \emph{learnable} spectral filter into low- and high-frequency
bands; a moving average is one particular fixed low-pass filter, and \method\
generalizes it by learning the band split end-to-end and giving each band its
own lightweight head. Against RLinear, \method\ shares the RevIN+linear backbone
but (a) replaces the single linear map with a frequency-decomposed set of heads
and (b) replaces vanilla RevIN with the horizon-adaptive \arevin\
(Section~\ref{sec:method:arevin}). The ablation that drops the frequency split
and closes the \arevin\ gate recovers an RLinear-like model, the controlled
comparison that isolates our contributions
(Section~\ref{sec:ablation}).

\subsection{Non-stationarity and Reversible Normalization}
\label{sec:related:norm}

Real series are non-stationary: mean and variance drift between the lookback and
the horizon, and between train and test. RevIN \citep{kim2022revin} is a
symmetric, instance-wise normalize-then-denormalize wrapper with a learnable
affine transform: it removes each instance's mean and standard deviation at the
input and restores them at the output, and it is now a near-universal plug-in.
Non-stationary Transformers \citep{liu2022nonstationary} make a subtler point:
naive stationarization can \emph{over-stationarize} and erase the very signal a
model needs, so they pair series stationarization with a de-stationary attention
that re-injects the removed statistics into the attention computation.

This is \method's clearest novelty. Vanilla RevIN applies a single,
horizon-agnostic de-normalization, adding back exactly the lookback mean and
variance to \emph{every} forecast step; but under distribution shift the
appropriate level and scale typically drift across the horizon, so the lookback
statistics are accurate for the first step and progressively worse for the last.
Our \arevin\ keeps RevIN's reversible structure but makes the de-normalization
horizon-aware and gated: a small learnable module modulates the restored
statistics as a function of horizon position, and a gate interpolates between
trusting the lookback statistics and letting the prediction set its own level.
This is philosophically aligned with the de-stationary idea of Non-stationary
Transformers---removed statistics should be re-injected \emph{adaptively} rather
than identically---but is realized as a lightweight, attention-free,
RevIN-compatible module suited to a linear backbone. To our knowledge, a
horizon-adaptive or gated RevIN of this form has not been published, and it is
the component least anticipated by prior linear or frequency-domain models
(Section~\ref{sec:related:positioning}).

\subsection{Frequency-Domain Forecasting}
\label{sec:related:freq}

A large body of work moves part or all of the forecasting computation into a
spectral representation, motivated by the fact that long-term series are
dominated by a few sparse, well-separated periodic components. Autoformer
\citep{wu2021autoformer} introduces the trend/seasonal series-decomposition
block (the moving-average split DLinear later reused) and replaces self-attention
with an Auto-Correlation mechanism that discovers period-based dependencies via
the FFT. FEDformer \citep{zhou2022fedformer} combines series decomposition with
frequency-enhanced attention over a random subset of Fourier (or wavelet) modes,
giving linear complexity and an explicit low-rank spectral filter. FiLM
\citep{zhou2022film} uses Legendre-polynomial projections to compress history and
Fourier projections to denoise it. TimesNet \citep{wu2023timesnet} uses the FFT
to find dominant periods, folds the 1D series into period-indexed 2D tensors, and
applies 2D convolutions. FreTS \citep{yi2023frets} is the most relevant of this
group: it discards attention entirely and learns frequency-domain MLPs over the
real and imaginary parts of the spectrum, across both channel and time
dimensions.

\method\ shares these models' premise---the frequency domain is the right place
to separate predictable structure from noise---but rejects their machinery.
FEDformer, FiLM, and TimesNet are heavy (attention, polynomial bases, or 2D
convolutions), with parameter and FLOP budgets far beyond the small-model regime
a 4\,GB GPU comfortably supports; FreTS is lighter but still uses complex-valued
MLPs across channels and time. \method\ instead uses the spectrum only to
\emph{split} the signal into bands, then forecasts each band with a plain
real-valued linear head in the time domain. The frequency domain is a routing
device, not the space in which the forecast is computed, which keeps the model
linear-scale while still exploiting the spectral separability these works
demonstrate.

\subsection{Transformer, MLP-Mixer, and Multiscale Forecasters}
\label{sec:related:transformer}

At the high-accuracy, high-cost end of the design space, Informer
\citep{zhou2021informer} makes long-sequence forecasting tractable with
ProbSparse attention and a generative decoder; PatchTST \citep{nie2023patchtst}
tokenizes the series into subseries patches and is strictly channel-independent,
and is the strongest Transformer forecaster on the standard benchmarks and our
small-Transformer baseline where memory permits; iTransformer
\citep{liu2024itransformer} inverts the tokenization so each variate becomes a
token and attention models cross-channel dependencies. These motivate the
central question of this paper: how much of their accuracy can a linear-scale
model recover at a fraction of the cost? We adopt PatchTST's channel-independent
treatment but reject patching and attention.

A parallel line removes attention while keeping expressive mixing. TSMixer
\citep{chen2023tsmixer} is an all-MLP architecture alternating time-mixing and
feature-mixing MLPs; TimeMixer \citep{wang2024timemixer} decomposes the series at
multiple sampling scales and mixes the disentangled multiscale components. Both
confirm that decomposition plus simple mixing is a strong, attention-free recipe,
and TimeMixer's multiscale decomposition is a temporal analogue of our spectral
band split. They differ from \method\ in weight class: they stack several mixing
blocks across scales and channels, whereas \method\ uses a single frequency split
with $K$ linear heads and no inter-channel mixing, targeting the smallest viable
model. The efficiency prong is grounded in the Green~AI agenda
\citep{schwartz2020greenai} and the energy-cost analysis of large models
\citep{strubell2019energy}, which argue that efficiency should be reported
alongside accuracy; \method\ reports parameters, FLOPs, train time, and peak
memory as first-class metrics. The Transformer \citep{vaswani2017attention} and
Adam \citep{kingma2015adam} are cited as foundational references.

\subsection{Positioning and Novelty}
\label{sec:related:positioning}

Two recent works are close enough that we position against them explicitly and
honestly. FITS \citep{xu2024fits} is the closest lightweight frequency model: it
applies rFFT to the lookback, a single complex-valued linear layer performing
amplitude scaling and phase shifting, then irFFT, using RevIN and a \emph{fixed
low-pass filter} that \emph{discards} high-frequency components (the source of
its $\sim$10k-parameter footprint), and is channel-independent. The overlap with
\method\ is real---both are RevIN + frequency-domain + lightweight + CI---and
reviewers will rightly raise it. \method\ differs from FITS in three concrete,
testable ways: (i) high-frequency content is \emph{modeled} by a dedicated head
rather than discarded by a low-pass filter, so residual structure that a low-pass
model leaves on the table can be exploited; (ii) the forecast is computed in the
time domain by $K$ real linear heads, not by a single complex map in the
frequency domain---a simpler hypothesis class; and (iii) the normalization is the
horizon-adaptive, gated \arevin, which has no analogue in FITS's plain RevIN.
Accordingly, we do \emph{not} claim to be the first lightweight frequency-domain
linear model---FITS holds that ground---and we include FITS as a run-in-budget
baseline.

FreqMoE \citep{liu2025freqmoe} also decomposes the spectrum into bands and
assigns each to an expert, with a gating network, learnable band boundaries, and
complex-valued residual blocks, using plain instance normalization. It shares the
``each band gets its own head'' idea but is substantially heavier---a
mixture-of-experts with complex-valued multi-block stacks---whereas \method\ uses
$K$ real linear heads recombined by addition, and FreqMoE has no adaptive
normalization contribution. As a 2025 arXiv-only work it is contemporaneous
rather than established prior art; we cite it and differentiate on weight class
and on \arevin. Taken together, the frequency band split with per-band heads is
not unique, so we present it as a specific lightweight design point rather than a
new concept, and we foreground \arevin---the prong with the least prior-art
overlap---as the primary methodological novelty.

\section{Method}
\label{sec:method}

\method\ is a channel-independent linear forecaster built from three components:
(i) a learnable, lossless frequency decomposition that routes the input into $K$
bands (Section~\ref{sec:method:freq}); (ii) one lightweight linear head per band
followed by recombination (Section~\ref{sec:method:heads}); and (iii) Adaptive
Reversible Instance Normalization (\arevin), a \emph{regime-adaptive} reversible
normalization that strictly generalizes RevIN \citep{kim2022revin}
(Section~\ref{sec:method:arevin}). Throughout, the design is deliberately
minimal: every added component is an opt-in generalization of a known linear
baseline, and each degenerates back to that baseline when its parameters take
their identity values. This yields the strict nesting
$\text{Linear} \subseteq \text{RLinear} \subseteq \method$, which forms the
backbone of the ablation study (Section~\ref{sec:ablation}).

\subsection{Problem Setup and Notation}
\label{sec:method:setup}

We observe a multivariate lookback window
$\bX \in \R^{B \times L \times C}$ of length $L$ over $C$ variates (channels)
and predict the future horizon $\bY \in \R^{B \times H \times C}$ of length
$H$, where $B$ is the batch size. Following the strong-baseline convention of
DLinear \citep{zeng2023dlinear}, RLinear \citep{li2023rlinear} and PatchTST
\citep{nie2023patchtst}, \method\ is \emph{channel-independent} (CI): every
variate is processed by the \emph{same} shared weights, and the channel
dimension is folded into the batch. Concretely, we reshape
$\bX : B{\times}L{\times}C \to (B{\cdot}C){\times}L$, operate on
$N = B{\cdot}C$ univariate series of length $L$, and reshape predictions back
to $B{\times}H{\times}C$. CI keeps the parameter count independent of $C$,
which is essential for the 4\,GB memory budget and for high-dimensional
datasets such as Electricity ($C{=}321$). Unless a tensor shape indicates
otherwise, the equations below are written for a single univariate series
$\bs \in \R^{L}$.

We use the real FFT throughout: \texttt{rfft}/\texttt{irfft} along the time
axis. For a real length-$L$ signal, \texttt{rfft} returns
$F = \lfloor L/2 \rfloor + 1$ non-redundant complex bins, and
$\texttt{irfft}(\cdot, n{=}L)$ inverts it exactly. All transforms are of length
$L$ (the lookback); the model never applies an FFT to the horizon. The default
and primary number of bands is $K{=}2$ (a low and a high band); $K \in \{3,4\}$
is explored only as an ablation.

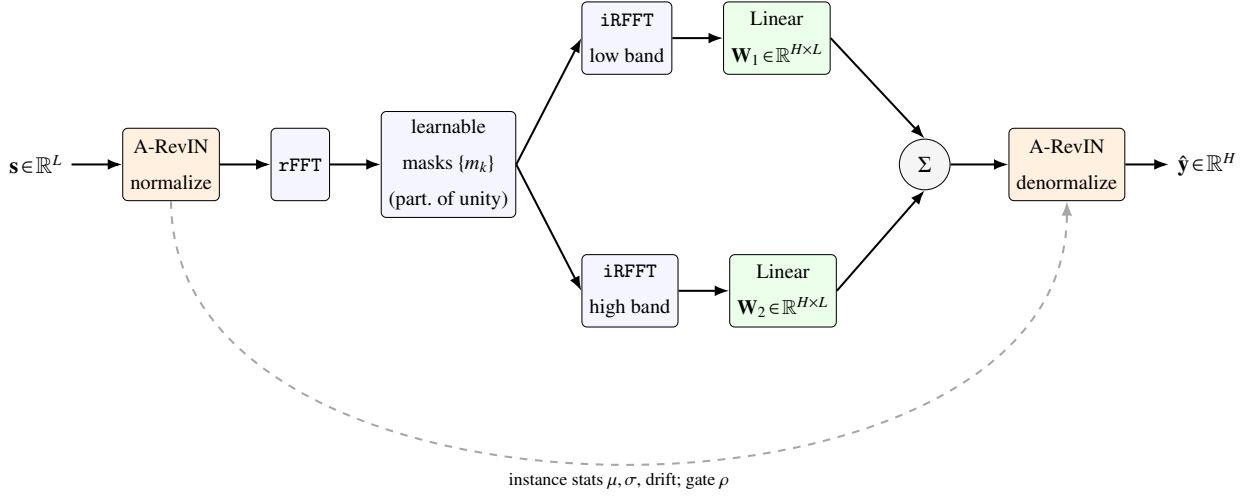
\begin{figure}[t]
  \centering
  \resizebox{\linewidth}{!}{%
  \begin{tikzpicture}[
    >={Latex[length=2mm]},
    node distance=6mm and 7mm,
    box/.style={draw, rounded corners=2pt, minimum height=10mm, align=center,
                font=\footnotesize, inner sep=3pt, fill=blue!4},
    norm/.style={box, fill=orange!12},
    head/.style={box, fill=green!8},
    op/.style={draw, circle, minimum size=7mm, font=\small, inner sep=0pt, fill=gray!8},
    io/.style={font=\small\itshape},
    arr/.style={->, thick},
  ]
    \node[io] (in) {$\bs\!\in\!\R^{L}$};
    \node[norm, right=of in]   (nrm)  {A-RevIN\\normalize};
    \node[box,  right=of nrm]  (fft)  {\texttt{rFFT}};
    \node[box,  right=of fft]  (mask) {learnable\\masks $\{m_k\}$\\(part.\ of unity)};
    \node[box,  above right=5mm and 9mm of mask] (irL)  {\texttt{iRFFT}\\low band};
    \node[box,  below right=5mm and 9mm of mask] (irH)  {\texttt{iRFFT}\\high band};
    \node[head, right=of irL]  (linL) {Linear\\$\mathbf{W}_1\!\in\!\R^{H\times L}$};
    \node[head, right=of irH]  (linH) {Linear\\$\mathbf{W}_2\!\in\!\R^{H\times L}$};
    \node[op] (sum) at ($(linL)!0.5!(linH)+(2.0cm,0)$) {$\Sigma$};
    \node[norm, right=8mm of sum] (dn) {A-RevIN\\denormalize};
    \node[io, right=6mm of dn] (out) {$\hat{\by}\!\in\!\R^{H}$};
    \draw[arr] (in) -- (nrm);
    \draw[arr] (nrm) -- (fft);
    \draw[arr] (fft) -- (mask);
    \draw[arr] (mask.east) -- (irL.west);
    \draw[arr] (mask.east) -- (irH.west);
    \draw[arr] (irL) -- (linL);
    \draw[arr] (irH) -- (linH);
    \draw[arr] (linL.east) -- (sum.north);
    \draw[arr] (linH.east) -- (sum.south);
    \draw[arr] (sum) -- (dn);
    \draw[arr] (dn) -- (out);
    \draw[->, thick, dashed, gray!70]
      (nrm.south) to[out=-90,in=-90]
      node[below, font=\scriptsize, text=black, midway]
      {instance stats $\mu,\sigma$, drift; gate $\rho$}
      (dn.south);
  \end{tikzpicture}%
  }
  \caption{Overview of the \method\ forward pass for one channel-independent
  series. A-RevIN normalizes the input and stores its instance statistics; a
  learnable partition-of-unity spectral filter splits the normalized series into
  $K$ bands; each band is forecast by a dedicated linear head; the band
  forecasts are summed and de-normalized by the horizon-adaptive A-RevIN
  inverse.}
  \label{fig:arch}
\end{figure}

Figure~\ref{fig:arch} gives the end-to-end forward pass: \arevin\ normalization,
the learnable frequency split, per-band linear heads, additive recombination,
and \arevin\ denormalization.

\subsection{Adaptive Reversible Instance Normalization (\arevin)}
\label{sec:method:arevin}

RevIN \citep{kim2022revin} normalizes each instance by its own lookback
mean/standard deviation, applies an optional affine transform, runs the model,
and then inverts the affine and re-applies the stored statistics. Its known
weakness is that it de-normalizes the \emph{entire} horizon with the
\emph{single} lookback statistic pair $(\mu, \sigma)$: under non-stationarity,
the horizon's true level and scale drift away from the lookback statistics, and
the error grows with horizon distance. \arevin\ retains RevIN's reversible
structure but makes the de-normalization \emph{horizon-adaptive} and
\emph{gated}, so that RevIN is recovered exactly as a special case.

\paragraph{Normalization (identical to RevIN).}
For a series $\bs \in \R^{L}$, statistics are taken over the time axis:
\begin{align}
\mu &= \frac{1}{L}\sum_{i=1}^{L} s_i, &
\sigma &= \sqrt{\tfrac{1}{L}\sum_{i=1}^{L}(s_i - \mu)^2 + \epsilon}, \\
\tilde{s}_i &= \frac{s_i - \mu}{\sigma}, &
s^{\mathrm{n}}_i &= \gamma\,\tilde{s}_i + \beta,
\label{eq:revin-norm}
\end{align}
with stabilizer $\epsilon = 10^{-5}$. The affine parameters
$\gamma, \beta \in \R$ are \emph{scalars} shared across all series and channels;
this preserves channel independence and keeps the parameter count independent of
$C$ (standard RevIN uses per-channel affine vectors of size $C$).

\paragraph{Adaptive de-normalization.}
Let $\bp \in \R^{H}$ be the band-recombined prediction in the normalized space
(Section~\ref{sec:method:heads}). Standard RevIN inverts every horizon step
$t \in \{1,\dots,H\}$ identically as
$\hat{y}_t = \sigma\,(p_t - \beta)/\gamma + \mu$. \arevin\ introduces three
learnable per-step vectors and one gate:
\begin{equation}
\mathbf{a} \in \R^{H},\quad
\mathbf{b} \in \R^{H},\quad
\boldsymbol{\lambda} \in \R^{H},\quad
\rho = \varsigma(r) \in (0,1),
\label{eq:arevin-params}
\end{equation}
where $\mathbf{a}$ is a per-step log-scale correction (initialized
$\mathbf{a}{=}\mathbf{0}$, so $\exp(0){=}1$), $\mathbf{b}$ is a per-step shift
correction in units of $\sigma$ (initialized $\mathbf{0}$),
$\boldsymbol{\lambda}$ is a per-step drift-propagation coefficient (initialized
$\mathbf{0}$), $\varsigma$ is the logistic sigmoid, and $\rho$ is a scalar gate
obtained from a raw parameter $r$. The initialization of $r$ is consequential and
is discussed below. We also compute a non-learnable \emph{drift} feature from
detached lookback statistics, capturing the recent-versus-early level change in
units of $\sigma$:
\begin{equation}
\mathrm{drift} =
\frac{\operatorname{mean}\!\big(\bs_{L/2:L}\big)
      - \operatorname{mean}\!\big(\bs_{0:L/2}\big)}{\sigma}.
\label{eq:drift}
\end{equation}
The horizon-adaptive inverse, applied per step $t = 1,\dots,H$, is
\begin{align}
\mathrm{scale}_t &= \exp\!\big(\rho\, a_t\big), \label{eq:arevin-scale}\\
\mathrm{shift}_t &= \rho\,\big(b_t\,\sigma + \lambda_t\,\mathrm{drift}\,\sigma\big),
  \label{eq:arevin-shift}\\
\hat{y}_t &= \mathrm{scale}_t \cdot
  \Big[\sigma\,\frac{p_t - \beta}{\gamma}\Big] + \mu + \mathrm{shift}_t.
  \label{eq:arevin-denorm}
\end{align}
The $\lambda_t$ term lets the model propagate the \emph{observed} lookback drift
into the horizon proportionally to step distance, providing a horizon-aware
extrapolation of level: $\lambda_t$ is expected to grow with $t$ on trending
datasets and to stay near zero on stationary ones.

\paragraph{Reversibility and identity guarantee.}
When $\rho = 0$ (equivalently
$\mathbf{a}=\mathbf{b}=\boldsymbol{\lambda}=\mathbf{0}$),
Equations~\eqref{eq:arevin-scale}--\eqref{eq:arevin-denorm} reduce \emph{exactly}
to the RevIN inverse $\hat{y}_t = \sigma\,(p_t-\beta)/\gamma + \mu$. Hence
$\text{RevIN} \subset \arevin$: training cannot do worse than RevIN at the
normalization layer up to optimization, and the gate $\rho$ lets the model learn
to stay at RevIN when adaptation does not help---making \arevin\
\emph{regime-adaptive}: it can engage where adaptation helps and reduce to RevIN
where it does not. \arevin\ adds $3H+3$ parameters
($\gamma,\beta,r$ plus $\mathbf{a},\mathbf{b},\boldsymbol{\lambda}$); for
$H{=}720$ this is $2163$ parameters, all shared across channels.

\paragraph{Gate initialization and a gradient trap.}
The reversibility guarantee makes a small gate value at initialization tempting,
but it conceals a gradient trap. Because the adaptive parameters start at their
identity values ($\mathbf{a}=\mathbf{b}=\boldsymbol{\lambda}=\mathbf{0}$), the
per-step corrections in
Equations~\eqref{eq:arevin-scale}--\eqref{eq:arevin-shift} are exactly zero at
initialization regardless of $\rho$, so $\partial\mathcal{L}/\partial\rho \approx
0$ and the gate receives essentially no gradient signal. Initializing $r$ so that
$\rho \approx 0$ therefore leaves \arevin\ dormant at RevIN throughout training,
even on data where adaptation would help. We instead initialize $r{=}0$ (so
$\rho{=}0.5$), which gives the gate a non-zero gradient and lets it open or close
as the data dictates; this is our default. Note that RevIN is still recovered
\emph{exactly} at the learned value $\rho{=}0$, so this initialization does not
compromise the identity guarantee---it only ensures the gate is trainable. We
quantify the effect of this choice in Section~\ref{sec:results:nonstat}.

\paragraph{Scope.}
\arevin\ is a \emph{first-order}, horizon-aware level/scale correction: it does
not model full distributional shift (there is no per-step variance modeling
beyond the scalar $\exp(\rho\,a_t)$), and the drift feature in
Equation~\eqref{eq:drift} is a simple two-half mean difference rather than a
learned encoder. This is deliberate, keeping \arevin\ at linear-model cost and
interpretable, and positioning it as a lightweight counterpart to heavier
stationarity-aware methods such as Non-stationary Transformers
\citep{liu2022nonstationary}.

\subsection{Learnable Frequency Decomposition}
\label{sec:method:freq}

DLinear \citep{zeng2023dlinear} decomposes a series into trend and seasonal
components using a \emph{fixed} moving-average kernel, i.e. a fixed-width
low-pass / high-pass split in the time domain. \method\ makes this split
\emph{learnable in the frequency domain} with a smooth, monotone,
parameter-light soft mask, and generalizes it from two bands to $K$. This is a
modest, defensible generalization of DLinear's decomposition; we explicitly do
\emph{not} claim to be the first frequency-domain linear forecaster
(cf. FEDformer \citep{zhou2022fedformer}, FiLM \citep{zhou2022film}, FreTS
\citep{yi2023frets}, and especially FITS \citep{xu2024fits}).

\paragraph{Soft spectral masks.}
Let the normalized frequency of rfft bin $f \in \{0,\dots,F-1\}$ be
\begin{equation}
\omega_f = \frac{f}{F-1} \in [0,1],
\qquad \omega_0 = 0\ (\text{DC}),\quad \omega_{F-1} = 1\ (\text{Nyquist}).
\end{equation}
For the primary case $K{=}2$ we learn a raw cutoff $c \in \R$ and a raw
sharpness $\tau \in \R$, and form
\begin{align}
\mathrm{cutoff} &= \varsigma(c) \in (0,1), &
\mathrm{sharpness} &= \operatorname{softplus}(\tau) + \epsilon_m \in (0,\infty),
\end{align}
with $\epsilon_m = 10^{-3}$, and define complementary low- and high-pass masks
\begin{align}
m_{\mathrm{low}}(\omega_f) &= \varsigma\!\big(-\,\mathrm{sharpness}\cdot(\omega_f - \mathrm{cutoff})\big),
  \label{eq:mask-low}\\
m_{\mathrm{high}}(\omega_f) &= 1 - m_{\mathrm{low}}(\omega_f).
  \label{eq:mask-high}
\end{align}
Here $m_{\mathrm{low}}$ is a smooth monotone-decreasing low-pass filter and
$m_{\mathrm{high}}$ is its exact complement, so the masks form a
\emph{partition of unity}: $\sum_k m_k(\omega_f) = 1$ for every bin. For general
$K$, we use $K-1$ ordered cutoffs $0 < c_1 < \dots < c_{K-1} < 1$ (ordering
enforced by a cumulative-softplus parameterization) and define
\begin{align}
m_1 &= \varsigma\!\big(-\tau_1(\omega - c_1)\big), \nonumber\\
m_k &= \varsigma\!\big(-\tau_k(\omega - c_k)\big)
       - \varsigma\!\big(-\tau_{k-1}(\omega - c_{k-1})\big),
       \quad 1 < k < K, \label{eq:mask-general}\\
m_K &= 1 - \varsigma\!\big(-\tau_{K-1}(\omega - c_{K-1})\big), \nonumber
\end{align}
which again partition unity by construction.

\paragraph{Application to the spectrum.}
The masks $m_k \in \R^{F}$ are real-valued and multiply the complex spectrum
bin-wise (the same real scalar scales the real and imaginary parts, so phase is
preserved and the gating is pure magnitude):
\begin{align}
\mathbf{S} &= \texttt{rfft}(\bs^{\mathrm{n}}, n{=}L), &
\mathbf{S}^{(k)} &= m_k \odot \mathbf{S}, &
\bs^{\mathrm{n},(k)} &= \texttt{irfft}\big(\mathbf{S}^{(k)}, n{=}L\big),
\label{eq:band-split}
\end{align}
where $\mathbf{S}, \mathbf{S}^{(k)} \in \mathbb{C}^{F}$ and
$\bs^{\mathrm{n},(k)} \in \R^{L}$. The masks depend only on the learnable scalars
(recomputed cheaply each forward pass) and are independent of both the batch and
the channel.

\paragraph{Properties.}
The decomposition has four properties that make it defensible:
(1)~\emph{Lossless partition}: since $\sum_k m_k = 1$ exactly,
$\sum_k \bs^{\mathrm{n},(k)} = \texttt{irfft}\big(\sum_k m_k \odot \mathbf{S}\big)
= \texttt{irfft}(\mathbf{S}) = \bs^{\mathrm{n}}$, so reconstruction is exact
\emph{before} the heads.
(2)~\emph{DLinear as a special case}: with $K{=}2$ and
$\mathrm{sharpness}\to\infty$, the soft mask approaches an ideal low/high split,
recovering the trend/residual inductive bias of DLinear's fixed box filter as one
member of a strictly richer, learnable family.
(3)~\emph{Differentiability}: $\mathrm{cutoff}$ and $\mathrm{sharpness}$ receive
gradients through \texttt{irfft}, so the split adapts to each dataset's spectral
content during training.
(4)~\emph{Cheapness}: the decomposition adds only $2(K-1)$ scalar parameters,
and the FFT/iFFT cost is $O(L\log L)$ per series, negligible relative to the
$H{\cdot}L$ linear heads.

\paragraph{Contrast with FITS.}
FITS \citep{xu2024fits}, the closest prior art, applies a low-pass cutoff in the
rFFT domain and forecasts with a single complex linear layer, thereby
\emph{discarding} all high-frequency content above the cutoff. \method's split is
different in kind: it is a partition of unity that \emph{keeps} all frequency
content and routes each band to its own dedicated head, so the high-frequency
band is \emph{modeled} rather than thrown away. This is a falsifiable difference
that the ablation in Section~\ref{sec:ablation} tests directly; if the spectral
split adds little over RLinear once \arevin\ is present, we report that plainly.

\paragraph{Initialization.}
For $K{=}2$ we initialize $\mathrm{cutoff}\approx 0.25$
(raw $c = \operatorname{logit}(0.25) \approx -1.0986$), so the low band starts as
roughly the lowest quarter of frequencies (a trend / low-season prior), and
$\mathrm{sharpness}\approx 10$. These are starting points and are subsequently
learned; we assert $0.02 < \mathrm{cutoff} < 0.98$ at initialization to avoid a
degenerate all-pass or all-stop mask, and we report the final learned values per
dataset as an interpretability artifact.

\subsection{Per-Band Linear Heads and Recombination}
\label{sec:method:heads}

\paragraph{Heads.}
Each band is forecast by one independent linear head mapping the lookback length
$L$ to the horizon length $H$:
\begin{equation}
\bp^{(k)} = \mathbf{W}_k\, \bs^{\mathrm{n},(k)} + \mathbf{b}_k,
\qquad \mathbf{W}_k \in \R^{H \times L},\ \mathbf{b}_k \in \R^{H},
\label{eq:head}
\end{equation}
realized as a single \texttt{nn.Linear}$(L,H)$ with bias. Each head has
$H{\cdot}L + H$ parameters.

\paragraph{Recombination.}
The primary model uses an identity sum,
\begin{equation}
\bp = \sum_{k=1}^{K} \bp^{(k)} \in \R^{H}.
\label{eq:recomb}
\end{equation}
Because the bands sum to the input (lossless partition) and each head is linear,
the identity sum keeps the model in the same hypothesis class as a single linear
head while granting the ability to apply a \emph{different} linear map to
different frequency content---which is the entire point. In particular, with
$K{=}1$, an all-pass mask, and identity recombination, \method\ degenerates
exactly to RLinear (RevIN + Linear); this nesting underpins the ablation table.
As an ablation we also consider a learnable band gate $\mathbf{g}\in\R^{K}$ with
$\bp = \sum_k g_k\,\bp^{(k)}$, reported only if it helps and we can explain why.

\subsection{Complexity and Parameter Count}
\label{sec:method:complexity}

For the primary configuration $K{=}2$, the total parameter count is
\begin{equation}
\underbrace{2\,(H{\cdot}L + H)}_{\text{heads}}
+ \underbrace{2(K-1)}_{\text{decomposition}}
+ \underbrace{3H + 3}_{\arevin}
\;\approx\; 2HL + 5H + 5.
\label{eq:params}
\end{equation}
For $L{=}96, H{=}96$ this is $\approx 18.9$k parameters, and for
$L{=}336, H{=}720$ it is $\approx 487.5$k. By comparison, DLinear uses
$\approx 2HL$ and RLinear $\approx HL$, so \method\ is roughly twice the size of
a single linear head---still linear-scale and easily within the 4\,GB budget. We
report this $2\times$ parameter cost honestly: the claim is competitive or better
accuracy at still-linear scale, not a free improvement. The per-forward compute
is dominated by the $K$ matrix--vector products of the heads ($O(KHL)$ per
series), with the FFT/iFFT adding only $O(L\log L)$; we report exact analytic
FLOPs alongside measured train time and peak memory in
Section~\ref{sec:efficiency}.

\section{Experimental Setup}
\label{sec:experiments}

A central goal of this work is to show that competitive long-term forecasting is
achievable under a strict commodity-hardware budget. Accordingly, every
experiment in this paper---training, evaluation, ablation, and efficiency
profiling, for \method\ and all baselines---is run on a \emph{single} NVIDIA
RTX~3050~Ti laptop GPU with 4\,GB of VRAM. We re-implement and run all baselines
ourselves under identical data splits, normalization, seeds, and hardware, so
all comparisons are like-for-like rather than copied from other papers.

\subsection{Datasets}
\label{sec:exp:data}

We evaluate on five widely used public long-term forecasting benchmarks: the
four Electricity Transformer Temperature datasets (ETTh1, ETTh2, ETTm1, ETTm2)
and Weather. The ETT datasets record seven power-system variates, with
\texttt{OT} (oil temperature) as the designated target; ETTh1/ETTh2 are sampled
hourly and ETTm1/ETTm2 every 15 minutes. Weather records 21 meteorological
variates at a 10-minute resolution. Table~\ref{tab:datasets} summarizes the
datasets. All series are multivariate and forecast in the channel-independent
setting (Section~\ref{sec:method:setup}). To probe behavior under distribution
shift, we additionally use two datasets outside the standard suite in the
non-stationarity study (Section~\ref{sec:results:nonstat}): Exchange-rate, which
is mildly non-stationary, and ILI (national illness), which is strongly
non-stationary and evaluated at the conventional shorter horizons
$H \in \{24,36,48,60\}$.

\begin{table}[t]
\centering
\caption{Summary of the benchmark datasets. ``Variates'' is the number of
channels $C$; ``Steps'' is the total length of the series. ETT splits follow the
standard 12/4/4-month train/validation/test convention; Weather uses a
70\%/10\%/20\% chronological split.}
\label{tab:datasets}
\begin{tabular}{lcccc}
\toprule
Dataset & Variates ($C$) & Frequency & Steps & Split \\
\midrule
ETTh1   & 7  & hourly   & 17{,}420 & 12/4/4 months \\
ETTh2   & 7  & hourly   & 17{,}420 & 12/4/4 months \\
ETTm1   & 7  & 15-min   & 69{,}680 & 12/4/4 months \\
ETTm2   & 7  & 15-min   & 69{,}680 & 12/4/4 months \\
Weather & 21 & 10-min   & 52{,}696 & 70/10/20 \% \\
\bottomrule
\end{tabular}
\end{table}

\subsection{Forecasting Protocol}
\label{sec:exp:protocol}

We follow the established long-term forecasting protocol
\citep{zeng2023dlinear,nie2023patchtst} so that our numbers are directly
comparable to published baselines. For the ETT datasets we use the canonical
12/4/4-month train/validation/test split; Weather uses a chronological
70\%/10\%/20\% split. Each series is z-score normalized using statistics
computed on the \emph{training portion only} (per channel), which prevents
leakage and matches the standard normalization that makes MSE/MAE comparable
across papers. The model's internal A-RevIN/RevIN then operates as an instance
normalization on top of this dataset-level normalization, and the loss and
metrics are computed in the train-z-scored space.

We report two lookback lengths: $L{=}336$, the strong setting for linear models,
as the \emph{headline} configuration, and $L{=}96$ for completeness. For every
lookback we forecast the four standard horizons $H \in \{96, 192, 336, 720\}$.
We evaluate with Mean Squared Error (MSE) and Mean Absolute Error (MAE),
averaged over all horizon steps and channels. Every $(\text{dataset}, H,
\text{model})$ cell is run with three seeds $\{2021, 2022, 2023\}$, and all
tables report the mean $\pm$ standard deviation over seeds.

\subsection{Baselines}
\label{sec:exp:baselines}

We compare against a sanity floor and five competitive forecasters spanning the
linear, lightweight-frequency, and Transformer families:
\begin{itemize}
  \item \textbf{Naive (repeat-last)}: predicts the last observed value for the
        entire horizon; a parameter-free sanity floor.
  \item \textbf{NLinear} \citep{zeng2023dlinear}: subtracts the last lookback
        value, applies a single \texttt{Linear}$(L,H)$, and adds it back---a
        minimal distribution-shift-robust linear model.
  \item \textbf{DLinear} \citep{zeng2023dlinear}: moving-average trend/seasonal
        decomposition with two \texttt{Linear}$(L,H)$ heads, summed; the
        time-domain decomposition baseline \method\ generalizes.
  \item \textbf{RLinear} \citep{li2023rlinear}: RevIN \citep{kim2022revin} plus a
        single \texttt{Linear}$(L,H)$; the normalization baseline \arevin\
        generalizes.
  \item \textbf{FITS} \citep{xu2024fits}: RevIN plus a low-pass cutoff in the
        rFFT domain and a single complex linear layer; the closest lightweight
        frequency-domain prior art, which \emph{discards} the high-frequency band
        \method\ retains.
  \item \textbf{PatchTST-small} \citep{nie2023patchtst}: a deliberately small,
        channel-independent patch Transformer (patch length 16, stride 8,
        $d_{\text{model}}{=}64$, 4 heads, 2 layers, dropout 0.2), included where
        it fits the 4\,GB budget as a strong Transformer reference.
\end{itemize}

\subsection{Implementation Details}
\label{sec:exp:impl}

All models are implemented in PyTorch 2.6.0 (CUDA 12.4) and trained on a single
RTX~3050~Ti laptop GPU (4\,GB). We use the Adam optimizer \citep{kingma2015adam}
with learning rate $10^{-3}$, no weight decay, and MSE loss. Training uses a
batch size of 32, at most 20 epochs with early stopping (patience 3 on the
validation loss), a type-1 step learning-rate schedule (halving the rate each
epoch after the first), and global gradient-norm clipping at 1.0. We train in
full \texttt{fp32} precision: the models are tiny, fp32 fits comfortably within
4\,GB, and it avoids the reproducibility noise of mixed precision. For
reproducibility we fix the \texttt{torch}, \texttt{numpy}, and \texttt{random}
seeds, enable deterministic algorithms, and disable cuDNN autotuning. At each
run we record parameters, analytic FLOPs, wall-clock training time per epoch, and
peak GPU memory via \texttt{torch.cuda.max\_memory\_allocated}; these populate the
efficiency analysis in Section~\ref{sec:efficiency}. Every reported number is
regenerable by a script in the repository.

\section{Results}
\label{sec:results}

We first present the main long-term forecasting comparison, then analyze
non-stationary regimes and the role of \arevin. All accuracy numbers are test
MSE and MAE in the train-standardized space, averaged over horizon steps and
channels and reported as mean (with standard deviation in the tables) over three
seeds; the best result in each row is shown in bold and the second best
underlined.

\subsection{Main Comparison}
\label{sec:results:main}

Table~\ref{tab:main_L336} reports the headline comparison at lookback $L{=}336$
across the five benchmarks and four horizons $H \in \{96,192,336,720\}$, against
the Naive floor, NLinear, DLinear, RLinear, FITS, and PatchTST-small;
Table~\ref{tab:main_L96} reports the corresponding $L{=}96$ results.

\IfFileExists{tables/main_table_L336.tex}{%
  \begingroup\fitwidetables
\begin{table*}[t]
\centering
\setlength{\tabcolsep}{3pt}
\caption{Long-term forecasting results (lookback $L=336$). Each cell is test MSE/MAE (mean$\pm$std over 3 seeds) in the train-standardized space. \textbf{Bold}: best; \underline{underline}: second best. PatchTST* uses a 4\,GB-feasible small config.}
\label{tab:main_L336}
\begin{tabular}{llcccccccccccccc}
\toprule
Dataset & $H$ & \multicolumn{2}{c}{Naive} & \multicolumn{2}{c}{NLinear} & \multicolumn{2}{c}{DLinear} & \multicolumn{2}{c}{RLinear} & \multicolumn{2}{c}{FITS} & \multicolumn{2}{c}{PatchTST*} & \multicolumn{2}{c}{FreqLite} \\
\cmidrule(lr){3-4} \cmidrule(lr){5-6} \cmidrule(lr){7-8} \cmidrule(lr){9-10} \cmidrule(lr){11-12} \cmidrule(lr){13-14} \cmidrule(lr){15-16}
& & MSE & MAE & MSE & MAE & MSE & MAE & MSE & MAE & MSE & MAE & MSE & MAE & MSE & MAE \\
\midrule
ETTh1 & 96 & 1.294\textpm0.000 & 0.713\textpm0.000 & 0.384\textpm0.005 & 0.405\textpm0.004 & \underline{0.376\textpm0.001} & \underline{0.398\textpm0.002} & 0.379\textpm0.003 & 0.400\textpm0.003 & 0.399\textpm0.000 & 0.419\textpm0.000 & 0.431\textpm0.043 & 0.434\textpm0.024 & \textbf{0.373\textpm0.001} & \textbf{0.395\textpm0.001} \\
 & 192 & 1.325\textpm0.000 & 0.733\textpm0.000 & 0.413\textpm0.001 & 0.421\textpm0.000 & 0.418\textpm0.009 & 0.428\textpm0.009 & \underline{0.412\textpm0.001} & \underline{0.419\textpm0.001} & 0.429\textpm0.000 & 0.436\textpm0.000 & 0.444\textpm0.006 & 0.445\textpm0.005 & \textbf{0.410\textpm0.005} & \textbf{0.417\textpm0.005} \\
 & 336 & 1.330\textpm0.000 & 0.746\textpm0.000 & \underline{0.438\textpm0.002} & \underline{0.437\textpm0.002} & 0.453\textpm0.011 & 0.454\textpm0.011 & 0.442\textpm0.007 & 0.440\textpm0.006 & 0.451\textpm0.000 & 0.448\textpm0.000 & 0.552\textpm0.058 & 0.503\textpm0.027 & \textbf{0.432\textpm0.001} & \textbf{0.430\textpm0.001} \\
 & 720 & 1.335\textpm0.000 & 0.755\textpm0.000 & \underline{0.444\textpm0.000} & \underline{0.459\textpm0.000} & 0.488\textpm0.016 & 0.501\textpm0.012 & 0.449\textpm0.002 & 0.463\textpm0.002 & 0.447\textpm0.001 & 0.466\textpm0.001 & 0.630\textpm0.029 & 0.557\textpm0.014 & \textbf{0.444\textpm0.002} & \textbf{0.459\textpm0.002} \\
\midrule
ETTh2 & 96 & 0.432\textpm0.000 & 0.422\textpm0.000 & 0.283\textpm0.001 & 0.344\textpm0.001 & 0.285\textpm0.003 & 0.350\textpm0.003 & \underline{0.281\textpm0.000} & \underline{0.342\textpm0.000} & 0.285\textpm0.001 & 0.346\textpm0.000 & 0.346\textpm0.016 & 0.382\textpm0.008 & \textbf{0.275\textpm0.001} & \textbf{0.336\textpm0.000} \\
 & 192 & 0.534\textpm0.000 & 0.473\textpm0.000 & 0.347\textpm0.003 & 0.386\textpm0.001 & 0.385\textpm0.009 & 0.420\textpm0.005 & 0.349\textpm0.013 & 0.386\textpm0.006 & \underline{0.344\textpm0.000} & \underline{0.383\textpm0.000} & 0.423\textpm0.029 & 0.429\textpm0.013 & \textbf{0.340\textpm0.005} & \textbf{0.380\textpm0.003} \\
 & 336 & 0.597\textpm0.000 & 0.511\textpm0.000 & 0.384\textpm0.013 & 0.415\textpm0.006 & 0.457\textpm0.007 & 0.469\textpm0.005 & 0.369\textpm0.004 & 0.407\textpm0.002 & \underline{0.364\textpm0.000} & \textbf{0.401\textpm0.000} & 0.432\textpm0.006 & 0.440\textpm0.003 & \textbf{0.362\textpm0.001} & \underline{0.402\textpm0.000} \\
 & 720 & 0.594\textpm0.000 & 0.519\textpm0.000 & 0.407\textpm0.007 & 0.442\textpm0.002 & 0.701\textpm0.027 & 0.594\textpm0.012 & 0.397\textpm0.001 & 0.433\textpm0.000 & \textbf{0.391\textpm0.000} & \textbf{0.427\textpm0.000} & 0.421\textpm0.016 & 0.447\textpm0.008 & \underline{0.392\textpm0.002} & \underline{0.430\textpm0.001} \\
\midrule
ETTm1 & 96 & 1.214\textpm0.000 & 0.665\textpm0.000 & 0.305\textpm0.002 & 0.347\textpm0.002 & \textbf{0.300\textpm0.001} & \underline{0.344\textpm0.000} & 0.305\textpm0.004 & 0.346\textpm0.003 & 0.304\textpm0.000 & 0.346\textpm0.000 & 0.302\textpm0.013 & 0.352\textpm0.008 & \underline{0.302\textpm0.001} & \textbf{0.344\textpm0.001} \\
 & 192 & 1.261\textpm0.000 & 0.690\textpm0.000 & 0.343\textpm0.003 & 0.370\textpm0.003 & \textbf{0.337\textpm0.002} & 0.368\textpm0.003 & 0.338\textpm0.003 & \underline{0.366\textpm0.003} & \underline{0.338\textpm0.000} & \textbf{0.366\textpm0.000} & 0.345\textpm0.001 & 0.381\textpm0.003 & 0.340\textpm0.004 & 0.367\textpm0.003 \\
 & 336 & 1.287\textpm0.000 & 0.707\textpm0.000 & 0.374\textpm0.002 & 0.386\textpm0.001 & \textbf{0.372\textpm0.001} & 0.390\textpm0.001 & 0.373\textpm0.001 & 0.386\textpm0.001 & 0.372\textpm0.000 & \underline{0.385\textpm0.000} & 0.381\textpm0.011 & 0.401\textpm0.006 & \underline{0.372\textpm0.001} & \textbf{0.384\textpm0.001} \\
 & 720 & 1.322\textpm0.000 & 0.729\textpm0.000 & 0.431\textpm0.003 & 0.419\textpm0.002 & \underline{0.428\textpm0.004} & 0.424\textpm0.006 & 0.431\textpm0.003 & 0.418\textpm0.001 & \textbf{0.428\textpm0.000} & \textbf{0.416\textpm0.000} & 0.429\textpm0.007 & 0.436\textpm0.006 & 0.428\textpm0.001 & \underline{0.417\textpm0.001} \\
\midrule
ETTm2 & 96 & 0.266\textpm0.000 & 0.328\textpm0.000 & 0.167\textpm0.001 & 0.257\textpm0.000 & 0.169\textpm0.002 & 0.265\textpm0.002 & \underline{0.166\textpm0.000} & \underline{0.255\textpm0.000} & 0.167\textpm0.000 & 0.256\textpm0.000 & 0.177\textpm0.002 & 0.262\textpm0.001 & \textbf{0.164\textpm0.001} & \textbf{0.253\textpm0.000} \\
 & 192 & 0.340\textpm0.000 & 0.371\textpm0.000 & 0.222\textpm0.000 & 0.294\textpm0.000 & 0.229\textpm0.006 & 0.308\textpm0.009 & \underline{0.221\textpm0.000} & \underline{0.292\textpm0.000} & 0.222\textpm0.000 & 0.293\textpm0.000 & 0.243\textpm0.009 & 0.308\textpm0.005 & \textbf{0.219\textpm0.000} & \textbf{0.290\textpm0.000} \\
 & 336 & 0.412\textpm0.000 & 0.410\textpm0.000 & 0.276\textpm0.000 & 0.329\textpm0.000 & 0.304\textpm0.023 & 0.362\textpm0.022 & \underline{0.275\textpm0.000} & 0.327\textpm0.000 & 0.275\textpm0.000 & \underline{0.327\textpm0.000} & 0.310\textpm0.009 & 0.350\textpm0.006 & \textbf{0.274\textpm0.001} & \textbf{0.326\textpm0.000} \\
 & 720 & 0.522\textpm0.000 & 0.466\textpm0.000 & 0.371\textpm0.000 & 0.386\textpm0.000 & 0.431\textpm0.017 & 0.443\textpm0.010 & \underline{0.369\textpm0.000} & 0.384\textpm0.000 & \textbf{0.368\textpm0.000} & \textbf{0.383\textpm0.000} & 0.387\textpm0.007 & 0.403\textpm0.007 & 0.370\textpm0.004 & \underline{0.383\textpm0.001} \\
\midrule
Weather & 96 & 0.259\textpm0.000 & 0.254\textpm0.000 & \underline{0.174\textpm0.001} & \underline{0.224\textpm0.002} & 0.174\textpm0.000 & 0.235\textpm0.001 & 0.175\textpm0.001 & 0.224\textpm0.001 & 0.176\textpm0.000 & 0.228\textpm0.000 & \textbf{0.149\textpm0.001} & \textbf{0.197\textpm0.001} & 0.174\textpm0.000 & 0.224\textpm0.001 \\
 & 192 & 0.309\textpm0.000 & 0.292\textpm0.000 & 0.217\textpm0.000 & 0.260\textpm0.000 & \underline{0.217\textpm0.001} & 0.275\textpm0.002 & 0.218\textpm0.000 & 0.260\textpm0.000 & 0.219\textpm0.000 & 0.263\textpm0.000 & \textbf{0.195\textpm0.002} & \textbf{0.241\textpm0.001} & 0.218\textpm0.000 & \underline{0.260\textpm0.001} \\
 & 336 & 0.376\textpm0.000 & 0.338\textpm0.000 & 0.266\textpm0.000 & 0.296\textpm0.000 & \underline{0.262\textpm0.001} & 0.313\textpm0.003 & 0.265\textpm0.000 & \underline{0.295\textpm0.000} & 0.267\textpm0.000 & 0.297\textpm0.000 & \textbf{0.248\textpm0.002} & \textbf{0.283\textpm0.004} & 0.265\textpm0.000 & 0.295\textpm0.000 \\
 & 720 & 0.465\textpm0.000 & 0.394\textpm0.000 & 0.334\textpm0.000 & 0.344\textpm0.000 & \underline{0.333\textpm0.007} & 0.376\textpm0.009 & 0.333\textpm0.000 & 0.342\textpm0.000 & 0.334\textpm0.000 & 0.343\textpm0.000 & \textbf{0.329\textpm0.008} & \textbf{0.339\textpm0.006} & 0.333\textpm0.000 & \underline{0.342\textpm0.000} \\
\bottomrule
\end{tabular}
\end{table*}
\endgroup%
}{\begin{table*}[t]\centering\caption{Main results at $L{=}336$.}%
  \label{tab:main_L336}\textit{[main results table at $L{=}336$ pending]}\end{table*}}

\IfFileExists{tables/main_table_L96.tex}{%
  \begingroup\fitwidetables
\begin{table*}[t]
\centering
\setlength{\tabcolsep}{3pt}
\caption{Long-term forecasting results (lookback $L=96$). Each cell is test MSE/MAE (mean$\pm$std over 3 seeds) in the train-standardized space. \textbf{Bold}: best; \underline{underline}: second best. PatchTST* uses a 4\,GB-feasible small config.}
\label{tab:main_L96}
\begin{tabular}{llcccccccccccccc}
\toprule
Dataset & $H$ & \multicolumn{2}{c}{Naive} & \multicolumn{2}{c}{NLinear} & \multicolumn{2}{c}{DLinear} & \multicolumn{2}{c}{RLinear} & \multicolumn{2}{c}{FITS} & \multicolumn{2}{c}{PatchTST*} & \multicolumn{2}{c}{FreqLite} \\
\cmidrule(lr){3-4} \cmidrule(lr){5-6} \cmidrule(lr){7-8} \cmidrule(lr){9-10} \cmidrule(lr){11-12} \cmidrule(lr){13-14} \cmidrule(lr){15-16}
& & MSE & MAE & MSE & MAE & MSE & MAE & MSE & MAE & MSE & MAE & MSE & MAE & MSE & MAE \\
\midrule
ETTh1 & 96 & 1.294\textpm0.000 & 0.713\textpm0.000 & 0.398\textpm0.003 & 0.407\textpm0.003 & 0.390\textpm0.003 & 0.404\textpm0.005 & 0.391\textpm0.000 & 0.399\textpm0.000 & 0.409\textpm0.001 & 0.417\textpm0.000 & \textbf{0.376\textpm0.003} & \underline{0.396\textpm0.000} & \underline{0.386\textpm0.000} & \textbf{0.394\textpm0.000} \\
 & 192 & 1.325\textpm0.000 & 0.733\textpm0.000 & 0.446\textpm0.001 & 0.433\textpm0.001 & \underline{0.440\textpm0.003} & 0.434\textpm0.004 & 0.443\textpm0.001 & \underline{0.429\textpm0.001} & 0.459\textpm0.001 & 0.446\textpm0.001 & 0.442\textpm0.000 & 0.436\textpm0.003 & \textbf{0.437\textpm0.000} & \textbf{0.423\textpm0.000} \\
 & 336 & 1.330\textpm0.000 & 0.746\textpm0.000 & 0.488\textpm0.001 & 0.454\textpm0.001 & 0.487\textpm0.004 & 0.464\textpm0.004 & \underline{0.486\textpm0.000} & \underline{0.450\textpm0.000} & 0.504\textpm0.005 & 0.470\textpm0.005 & 0.507\textpm0.003 & 0.469\textpm0.002 & \textbf{0.481\textpm0.000} & \textbf{0.446\textpm0.000} \\
 & 720 & 1.335\textpm0.000 & 0.755\textpm0.000 & \underline{0.482\textpm0.000} & \underline{0.472\textpm0.000} & 0.510\textpm0.002 & 0.505\textpm0.002 & 0.487\textpm0.000 & 0.474\textpm0.000 & 0.496\textpm0.007 & 0.488\textpm0.005 & 0.529\textpm0.022 & 0.500\textpm0.007 & \textbf{0.482\textpm0.000} & \textbf{0.470\textpm0.000} \\
\midrule
ETTh2 & 96 & 0.432\textpm0.000 & 0.422\textpm0.000 & 0.298\textpm0.000 & 0.347\textpm0.000 & 0.327\textpm0.002 & 0.382\textpm0.003 & \underline{0.295\textpm0.000} & \underline{0.344\textpm0.000} & 0.303\textpm0.000 & 0.349\textpm0.001 & 0.298\textpm0.007 & 0.346\textpm0.004 & \textbf{0.290\textpm0.000} & \textbf{0.339\textpm0.000} \\
 & 192 & 0.534\textpm0.000 & 0.473\textpm0.000 & 0.385\textpm0.000 & 0.399\textpm0.000 & 0.450\textpm0.005 & 0.457\textpm0.003 & 0.382\textpm0.000 & 0.396\textpm0.000 & 0.387\textpm0.001 & 0.398\textpm0.001 & \textbf{0.376\textpm0.006} & \underline{0.394\textpm0.004} & \underline{0.376\textpm0.000} & \textbf{0.391\textpm0.000} \\
 & 336 & 0.597\textpm0.000 & 0.511\textpm0.000 & 0.425\textpm0.001 & 0.434\textpm0.000 & 0.567\textpm0.010 & 0.527\textpm0.005 & \underline{0.421\textpm0.000} & \underline{0.431\textpm0.000} & 0.424\textpm0.001 & 0.432\textpm0.000 & 0.424\textpm0.001 & 0.432\textpm0.002 & \textbf{0.416\textpm0.000} & \textbf{0.427\textpm0.000} \\
 & 720 & 0.594\textpm0.000 & 0.519\textpm0.000 & 0.430\textpm0.000 & 0.450\textpm0.000 & 0.781\textpm0.027 & 0.635\textpm0.012 & 0.427\textpm0.000 & 0.445\textpm0.000 & 0.429\textpm0.000 & 0.444\textpm0.001 & \underline{0.425\textpm0.009} & \underline{0.443\textpm0.005} & \textbf{0.424\textpm0.002} & \textbf{0.442\textpm0.001} \\
\midrule
ETTm1 & 96 & 1.214\textpm0.000 & 0.665\textpm0.000 & 0.349\textpm0.002 & \underline{0.371\textpm0.002} & \underline{0.344\textpm0.002} & 0.372\textpm0.003 & 0.352\textpm0.001 & 0.373\textpm0.001 & 0.362\textpm0.000 & 0.381\textpm0.000 & \textbf{0.324\textpm0.004} & \textbf{0.364\textpm0.003} & 0.351\textpm0.002 & 0.372\textpm0.000 \\
 & 192 & 1.261\textpm0.000 & 0.690\textpm0.000 & 0.391\textpm0.000 & 0.391\textpm0.000 & \underline{0.381\textpm0.001} & 0.391\textpm0.000 & 0.391\textpm0.001 & 0.392\textpm0.001 & 0.397\textpm0.000 & 0.397\textpm0.000 & \textbf{0.369\textpm0.003} & \textbf{0.391\textpm0.002} & 0.389\textpm0.002 & \underline{0.391\textpm0.002} \\
 & 336 & 1.287\textpm0.000 & 0.707\textpm0.000 & 0.424\textpm0.002 & \underline{0.413\textpm0.001} & \underline{0.413\textpm0.000} & 0.413\textpm0.001 & 0.424\textpm0.001 & 0.414\textpm0.001 & 0.429\textpm0.000 & 0.417\textpm0.000 & \textbf{0.391\textpm0.001} & \textbf{0.407\textpm0.001} & 0.423\textpm0.001 & 0.413\textpm0.001 \\
 & 720 & 1.322\textpm0.000 & 0.729\textpm0.000 & 0.486\textpm0.000 & \underline{0.448\textpm0.000} & \underline{0.474\textpm0.003} & 0.453\textpm0.005 & 0.487\textpm0.000 & 0.449\textpm0.000 & 0.489\textpm0.000 & 0.450\textpm0.000 & \textbf{0.451\textpm0.002} & \textbf{0.442\textpm0.000} & 0.487\textpm0.002 & 0.449\textpm0.000 \\
\midrule
ETTm2 & 96 & 0.266\textpm0.000 & 0.328\textpm0.000 & 0.183\textpm0.000 & 0.266\textpm0.001 & 0.191\textpm0.002 & 0.287\textpm0.004 & 0.183\textpm0.000 & 0.266\textpm0.000 & 0.186\textpm0.000 & 0.269\textpm0.000 & \textbf{0.177\textpm0.001} & \textbf{0.261\textpm0.000} & \underline{0.182\textpm0.000} & \underline{0.264\textpm0.001} \\
 & 192 & 0.340\textpm0.000 & 0.371\textpm0.000 & 0.247\textpm0.000 & 0.305\textpm0.001 & 0.273\textpm0.007 & 0.348\textpm0.010 & 0.247\textpm0.000 & 0.306\textpm0.000 & 0.249\textpm0.000 & 0.307\textpm0.000 & \textbf{0.245\textpm0.003} & \textbf{0.304\textpm0.003} & \underline{0.246\textpm0.000} & \underline{0.304\textpm0.000} \\
 & 336 & 0.412\textpm0.000 & 0.410\textpm0.000 & 0.309\textpm0.000 & 0.344\textpm0.000 & 0.348\textpm0.015 & 0.400\textpm0.013 & \underline{0.308\textpm0.000} & 0.344\textpm0.000 & 0.309\textpm0.000 & \underline{0.344\textpm0.000} & 0.310\textpm0.005 & 0.347\textpm0.005 & \textbf{0.308\textpm0.000} & \textbf{0.343\textpm0.000} \\
 & 720 & 0.522\textpm0.000 & 0.466\textpm0.000 & 0.409\textpm0.000 & 0.400\textpm0.000 & 0.527\textpm0.053 & 0.505\textpm0.032 & 0.409\textpm0.000 & 0.399\textpm0.000 & 0.409\textpm0.000 & \underline{0.399\textpm0.000} & \textbf{0.407\textpm0.004} & 0.405\textpm0.003 & \underline{0.408\textpm0.000} & \textbf{0.398\textpm0.000} \\
\midrule
Weather & 96 & 0.259\textpm0.000 & 0.254\textpm0.000 & 0.195\textpm0.000 & 0.234\textpm0.000 & 0.197\textpm0.002 & 0.257\textpm0.004 & \underline{0.195\textpm0.000} & 0.234\textpm0.000 & 0.197\textpm0.000 & 0.238\textpm0.000 & \textbf{0.175\textpm0.001} & \textbf{0.217\textpm0.001} & 0.195\textpm0.000 & \underline{0.234\textpm0.000} \\
 & 192 & 0.309\textpm0.000 & 0.292\textpm0.000 & 0.241\textpm0.000 & 0.271\textpm0.000 & 0.244\textpm0.007 & 0.306\textpm0.010 & 0.240\textpm0.000 & 0.270\textpm0.000 & 0.242\textpm0.000 & 0.273\textpm0.000 & \textbf{0.221\textpm0.000} & \textbf{0.258\textpm0.000} & \underline{0.240\textpm0.000} & \underline{0.270\textpm0.000} \\
 & 336 & 0.376\textpm0.000 & 0.338\textpm0.000 & 0.293\textpm0.000 & 0.308\textpm0.000 & \underline{0.285\textpm0.002} & 0.336\textpm0.002 & 0.291\textpm0.000 & 0.306\textpm0.000 & 0.294\textpm0.000 & 0.309\textpm0.000 & \textbf{0.278\textpm0.001} & \textbf{0.299\textpm0.001} & 0.292\textpm0.000 & \underline{0.306\textpm0.000} \\
 & 720 & 0.465\textpm0.000 & 0.394\textpm0.000 & 0.366\textpm0.000 & 0.356\textpm0.000 & \textbf{0.350\textpm0.004} & 0.387\textpm0.006 & 0.364\textpm0.000 & \underline{0.353\textpm0.000} & 0.366\textpm0.000 & 0.355\textpm0.000 & \underline{0.355\textpm0.001} & \textbf{0.349\textpm0.001} & 0.364\textpm0.001 & 0.353\textpm0.000 \\
\bottomrule
\end{tabular}
\end{table*}
\endgroup%
}{\begin{table*}[t]\centering\caption{Main results at $L{=}96$.}%
  \label{tab:main_L96}\textit{[main results table at $L{=}96$ pending]}\end{table*}}

At the headline $L{=}336$ setting, \method\ attains the best average test MSE of
$0.3244$ over the $20$ dataset$\times$horizon cells, ahead of RLinear ($0.3273$),
NLinear ($0.3290$), and FITS ($0.3291$), and well ahead of DLinear ($0.3559$) and
the PatchTST-small Transformer ($0.3587$); the Naive floor is $0.7738$. \method\
wins the most cells outright ($10$ of $20$), versus PatchTST ($4$), FITS ($3$),
and DLinear ($3$). It is thus the strongest lightweight model on the standard
benchmarks and, notably, surpasses the Transformer baseline at long lookback
while using roughly $4\times$ fewer parameters and $2.2\times$ less memory
(Section~\ref{sec:efficiency}).

The value of long lookback for linear-scale models is underscored at $L{=}96$
(Table~\ref{tab:main_L96}), where PatchTST-small is the overall best ($0.3541$
average MSE, $12$ wins): with a short context the Transformer's representational
capacity pays off. Among the lightweight models, however, \method\ remains best
($0.3589$, $7$ wins), ahead of RLinear ($0.3611$), NLinear ($0.3623$), FITS
($0.3669$), and DLinear ($0.3989$). The headline regime for a linear-scale
forecaster is the long-lookback setting, and there \method\ leads all baselines
including the Transformer.

\subsection{Discussion}
\label{sec:results:discussion}

Two findings stand out. First, the accuracy gains of \method\ over the strongest
linear baseline (RLinear) on the standard, largely stationary benchmarks are
\emph{modest}: the average-MSE gap at $L{=}336$ is about $0.9\%$. The headline
strengths on these datasets are therefore (i) being the best lightweight model
overall and (ii) surpassing the PatchTST Transformer at long lookback at a small
fraction of its cost---an efficiency result we quantify in
Section~\ref{sec:efficiency}---rather than a large accuracy margin over RLinear.
We report this plainly. Crucially, though modest in magnitude, this gain over
RLinear is \emph{statistically robust rather than noise}: a paired Wilcoxon
signed-rank test across all $60$ $(\text{dataset},\text{horizon},\text{seed})$
cells gives $p \approx 1.2\times10^{-6}$ at $L{=}336$
(Table~\ref{tab:significance}), and \method\ improves significantly over every
linear and lightweight-frequency baseline at both lookbacks
($p < 10^{-3}$ in all such comparisons). The one comparison in which \method\
does not lead is against PatchTST at the short lookback $L{=}96$ ($-1.36\%$),
consistent with the Transformer's short-context advantage discussed above; at the
headline $L{=}336$ setting \method\ surpasses PatchTST by $9.57\%$
($p \approx 8.7\times10^{-6}$). Second, the picture changes on
\emph{non-stationary} data, where \method's regime-adaptive normalization engages
and yields substantially larger gains; we analyze this next.

\IfFileExists{tables/significance_table.tex}{%
\begin{table}[t]\centering
\caption{Paired significance tests for \method\ versus each baseline (Wilcoxon
signed-rank over all matched $(\text{dataset},\text{horizon},\text{seed})$
cells). $\Delta$MSE is the mean relative improvement of \method\ (positive =
\method\ better); $n$ is the number of paired cells.
$^{***}p<10^{-3}$, $^{*}p<0.05$.}
\label{tab:significance}
\begin{tabular}{llrrl}
\toprule
Setting & Baseline & $n$ & $\Delta$MSE (\%) & $p$ (Wilcoxon) \\
\midrule
$L{=}336$ & RLINEAR & 60 & +0.89 & 1.2e-06~(***) \\
$L{=}336$ & NLINEAR & 60 & +1.41 & 2.4e-08~(***) \\
$L{=}336$ & DLINEAR & 60 & +8.84 & 1.7e-06~(***) \\
$L{=}336$ & FITS & 60 & +1.42 & 5.6e-07~(***) \\
$L{=}336$ & PatchTST & 60 & +9.57 & 8.7e-06~(***) \\
$L{=}96$ & RLINEAR & 60 & +0.60 & 7.4e-08~(***) \\
$L{=}96$ & NLINEAR & 60 & +0.94 & 2.5e-08~(***) \\
$L{=}96$ & DLINEAR & 60 & +10.04 & 5.4e-04~(***) \\
$L{=}96$ & FITS & 60 & +2.17 & 1.7e-11~(***) \\
$L{=}96$ & PatchTST & 60 & -1.36 & 1.4e-02~(*) \\
ILI & RLINEAR & 12 & +3.90 & 4.9e-04~(***) \\
\bottomrule
\end{tabular}
\end{table}}{\begin{table}[t]\centering\caption{Significance tests.}%
\label{tab:significance}\textit{[significance table pending]}\end{table}}

\subsection{Non-stationary Regimes and the Role of \arevin}
\label{sec:results:nonstat}

\arevin\ is designed to engage under non-stationarity and to reduce to RevIN
otherwise. To test this directly we evaluate two datasets outside the standard
suite: Exchange-rate, which is only mildly non-stationary, and ILI (national
illness), which is strongly non-stationary. Table~\ref{tab:nonstationary}
reports MSE on both.

\IfFileExists{tables/nonstationary_table.tex}{%
\begin{table}[t]\centering
\caption{Non-stationarity study (test MSE; lower is better). Exchange-rate is
mildly non-stationary; ILI is strongly non-stationary. Best per row in
\textbf{bold}.}
\label{tab:nonstationary}
\begin{tabular}{llcccc}
\toprule
Dataset & $H$ & RLinear & DLinear & FITS & FreqLite \\
\midrule
Exchange & 96 & 0.084 & \textbf{0.081} & 0.098 & 0.084 \\
 & 192 & 0.175 & \textbf{0.165} & 0.193 & 0.175 \\
 & 336 & 0.323 & \textbf{0.299} & 0.344 & 0.324 \\
 & 720 & 0.836 & \textbf{0.801} & 0.884 & 0.843 \\
\midrule
ILI & 24 & 4.050 & 4.684 & 6.667 & \textbf{3.888} \\
 & 36 & 3.774 & 4.648 & 6.646 & \textbf{3.593} \\
 & 48 & 3.469 & 4.648 & 6.037 & \textbf{3.349} \\
 & 60 & 3.406 & 4.926 & 6.182 & \textbf{3.296} \\
\bottomrule
\end{tabular}
\end{table}}{\begin{table}[t]\centering\caption{Non-stationarity study.}%
\label{tab:nonstationary}\textit{[non-stationary table pending]}\end{table}}

On the mildly non-stationary Exchange-rate data, DLinear is the best model and
\arevin\ is neutral-to-slightly-negative: isolating its effect by comparing
RLinear with an A-RevIN-Linear of the same backbone gives differences of only
$+0.000$, $-0.000$, $-0.001$, and $-0.002$ across the four horizons. We report
honestly that \method\ is competitive \emph{within} the RevIN family here but
that DLinear's time-domain decomposition wins on this dataset.

On the strongly non-stationary ILI data, \arevin\ clearly helps. Isolating its
effect on the same single-head backbone (RLinear vs.\ A-RevIN-Linear), the
adaptive normalization reduces MSE by $0.66\%$, $0.63\%$, $0.52\%$, and $0.38\%$
at horizons $H \in \{24,36,48,60\}$. The full \method\ improves over RLinear by
$4.0\%$, $4.8\%$, $3.5\%$, and $3.2\%$ (mean $3.9\%$) across the same horizons,
while DLinear and FITS are far worse on ILI (MSE in the $4.6$--$6.6$ range versus
\method's $3.3$--$3.9$).

\paragraph{Gate sweep and a gradient-trap finding.}
Table~\ref{tab:arevin_gate} sweeps the \arevin\ gate on ILI. As the learned mean
gate $\bar\rho$ increases from $0.02$ (with raw gate initialization
$\rho_0{=}{-}4$) through $0.51$ ($\rho_0{=}0$) to $1.00$ (forced), the ILI error
decreases \emph{monotonically}. This exposes a methodological subtlety: with
$\rho_0{=}{-}4$ the gate is gradient-starved. Because the adaptive parameters are
initialized at their identity values
($\mathbf{a}=\mathbf{b}=\boldsymbol{\lambda}=\mathbf{0}$), the gradient
$\partial\mathcal{L}/\partial\rho$ is approximately zero at initialization, so the
gate never opens and \arevin\ stays at RevIN even on data where adapting would
help (the gate remained at $\bar\rho{=}0.02$ on ILI). Initializing $\rho_0{=}0$,
now our default, lets the gate engage, while RevIN is still recovered exactly at
$\rho{=}0$. This is the recommended initialization for \arevin.

\IfFileExists{tables/arevin_gate_table.tex}{%
\begin{table}[t]\centering
\caption{\arevin\ gate sweep on ILI (test MSE; lower is better). $\bar\rho$ is
the learned mean gate and $\rho_0$ the raw gate initialization. Increasing gate
engagement monotonically lowers error.}
\label{tab:arevin_gate}
\begin{tabular}{lcccccc}
\toprule
Model & $\bar\rho$ & $H{=}24$ & $H{=}36$ & $H{=}48$ & $H{=}60$ & Avg \\
\midrule
RLinear (RevIN) & -- & 4.050 & 3.774 & 3.469 & 3.406 & 3.675 \\
A-RevIN-Linear & 0.51 & 4.023 & 3.750 & 3.451 & 3.393 & 3.654 \\
FreqLite ($\rho_0{=}{-}4$, trapped) & 0.02 & 3.911 & 3.613 & 3.365 & 3.308 & 3.549 \\
FreqLite ($\rho_0{=}0$) & 0.51 & 3.888 & 3.593 & 3.349 & 3.296 & 3.531 \\
FreqLite (forced, $\rho{\approx}1$) & 1.00 & 3.866 & 3.575 & 3.334 & 3.285 & \textbf{3.515} \\
\bottomrule
\end{tabular}
\end{table}}{\begin{table}[t]\centering\caption{A-RevIN gate sweep on ILI.}%
\label{tab:arevin_gate}\textit{[gate-sweep table pending]}\end{table}}

\begin{figure}[t]
  \centering
  \IfFileExists{figures/learned_filter.pdf}{%
    \includegraphics[width=0.82\linewidth]{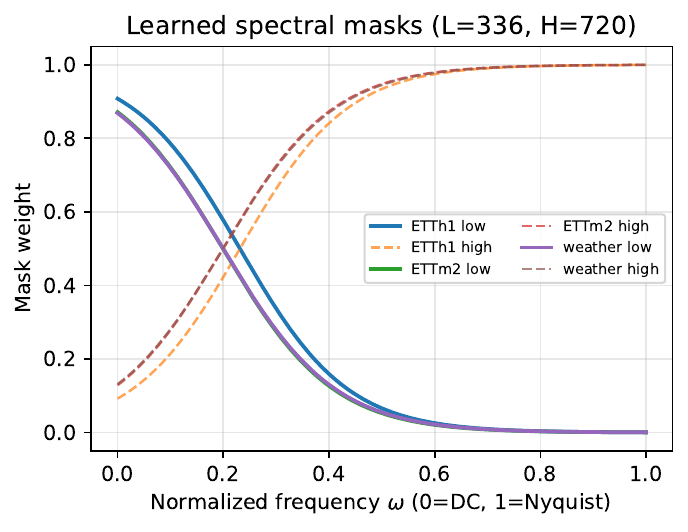}%
  }{\fbox{\parbox[c][0.20\textheight][c]{0.8\linewidth}{\centering
    \textbf{[learned filter figure pending]}}}}
  \caption{Learned low/high spectral masks of \method's partition-of-unity
  decomposition as a function of normalized frequency.}
  \label{fig:learned_filter}
\end{figure}

\begin{figure}[t]
  \centering
  \IfFileExists{figures/arevin_profile.pdf}{%
    \includegraphics[width=0.82\linewidth]{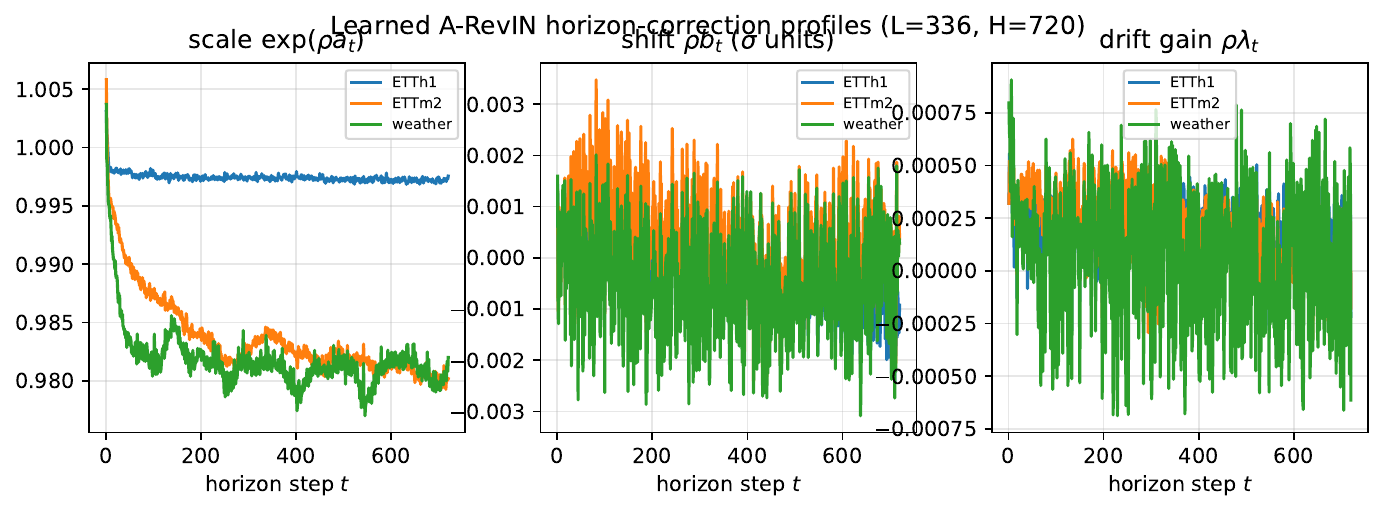}%
  }{\fbox{\parbox[c][0.20\textheight][c]{0.8\linewidth}{\centering
    \textbf{[A-RevIN profile figure pending]}}}}
  \caption{Per-step \arevin\ correction profile across the forecast horizon,
  showing how the adaptive de-normalization varies with horizon distance.}
  \label{fig:arevin_profile}
\end{figure}

\paragraph{Interpretability.}
Figure~\ref{fig:learned_filter} shows the learned spectral mask, and
Figure~\ref{fig:arevin_profile} the per-step \arevin\ correction profile, giving
direct evidence that the components behave as designed rather than merely adding
parameters.

\subsection{A Controlled Study: \arevin\ Engages with Non-stationarity}
\label{sec:results:synthetic}

The ILI result shows \arevin\ helping on a real strongly non-stationary dataset,
but real data confounds non-stationarity with many other factors. To confirm the
mechanism directly, we construct a synthetic benchmark in which the
non-stationarity magnitude is the \emph{only} thing that varies. Each series has
four channels and length $4000$ and is the sum of (i) a stationary seasonal base
with periods $24$ and $168$, (ii) Gaussian observation noise, and (iii)
$\delta$ times a unit-variance \emph{persistent} trend---an AR(1)-smoothed random
slope that is locally linear and therefore extrapolatable. The seasonal base and
the trend signal are \emph{shared} across all values of $\delta$, so increasing
$\delta$ injects more non-stationarity while holding everything else fixed; at
$\delta{=}0$ the series is stationary. We isolate \arevin\ by comparing RLinear
(plain RevIN) against an A-RevIN-Linear ($K{=}1$, identical backbone) at
$L{=}96, H{=}96$ over three seeds.

Table~\ref{tab:synthetic} reports the MSE reduction of \arevin\ over RevIN and the
learned mean gate $\bar\rho$ as $\delta$ increases. The benefit rises
\emph{monotonically} with injected drift---from neutral-to-slightly-negative when
the series is stationary or mildly non-stationary ($-0.28\%$ at $\delta{=}0$,
$-0.23\%$ at $\delta{=}0.5$, $-0.09\%$ at $\delta{=}1$) to a clear positive gain
as drift grows ($+0.30\%$ at $\delta{=}2$, $+1.42\%$ at $\delta{=}4$), with the
learned gate widening accordingly ($\bar\rho$ from $0.52$ to $0.54$). This is a
controlled confirmation of the mechanism that complements the ILI result: under
stationarity \arevin\ is harmless and reduces to RevIN, and its advantage grows
precisely with the non-stationarity it is designed to handle. As elsewhere, we
note honestly that the absolute gains are modest.

\IfFileExists{tables/synthetic_table.tex}{%
\begin{table}[t]\centering
\caption{Controlled synthetic-drift study at $L{=}96, H{=}96$ (3 seeds). $\delta$
scales the injected persistent trend ($\delta{=}0$ is stationary). $\Delta\%$ is
the MSE reduction of A-RevIN-Linear over RLinear (plain RevIN) on the identical
backbone; $\bar\rho$ is \arevin's learned mean gate. The benefit increases
monotonically with $\delta$.}
\label{tab:synthetic}
\begin{tabular}{lrrrr}
\toprule
$\delta$ & RLinear & A-RevIN-Lin. & $\Delta$\% & learned $\bar\rho$ \\
\midrule
0 & 0.1150 & 0.1153 & -0.28 & 0.52 \\
0.5 & 0.1227 & 0.1230 & -0.23 & 0.52 \\
1 & 0.1348 & 0.1349 & -0.09 & 0.52 \\
2 & 0.1382 & 0.1378 & +0.30 & 0.52 \\
4 & 0.1097 & 0.1081 & +1.42 & 0.54 \\
\bottomrule
\end{tabular}
\end{table}}{\begin{table}[t]\centering\caption{Synthetic-drift study.}%
\label{tab:synthetic}\textit{[synthetic table pending]}\end{table}}

\begin{figure}[t]
  \centering
  \IfFileExists{figures/synthetic_drift.pdf}{%
    \includegraphics[width=0.82\linewidth]{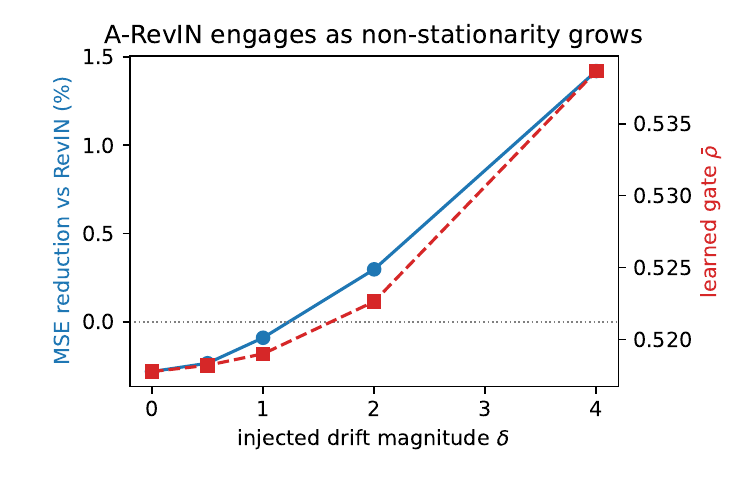}%
  }{\fbox{\parbox[c][0.20\textheight][c]{0.8\linewidth}{\centering
    \textbf{[synthetic-drift figure pending]}}}}
  \caption{\arevin's benefit and its learned gate both increase with injected
  non-stationarity: MSE reduction over RevIN and learned mean gate $\bar\rho$ as
  functions of the controlled drift magnitude $\delta$.}
  \label{fig:synthetic}
\end{figure}

\section{Ablation Study}
\label{sec:ablation}

We isolate the contribution of each \method\ component on a representative subset
of the standard (largely stationary) benchmarks---ETTh1, ETTm2, and
Weather---at horizons $H \in \{96, 720\}$ with lookback $L{=}336$ and three
seeds. Because every component is an opt-in generalization that degenerates to a
known baseline (Section~\ref{sec:method}), each ablation is a clean controlled
comparison along the nesting $\text{Linear} \subseteq \text{RLinear} \subseteq
\method$.

\subsection{Component Ablations}
\label{sec:ablation:components}

Table~\ref{tab:ablation} reports the ablations. We consider: (A0) the full
\method; (A1) replacing \arevin\ with plain RevIN while keeping the decomposition,
which isolates the value of adaptive de-normalization; (A2) replacing the
learnable spectral split with a DLinear-style fixed moving-average split while
keeping \arevin, which isolates the value of the learnable decomposition; (A3)
$K{=}1$ with \arevin\ (an ``A-RevIN-Linear''), showing the decomposition's
marginal value; (A4) $K{=}1$ with plain RevIN, which is exactly RLinear and
verifies the degeneracy $\method \supseteq \text{RLinear}$; (A5) increasing the
number of bands to $K \in \{3,4\}$; (A7) freezing the decomposition at
initialization to separate the benefit of a soft mask from that of
\emph{learning} it; and (A8) \arevin\ without the drift-propagation term
$\boldsymbol{\lambda}$.

\IfFileExists{tables/ablation_table.tex}{%
\begin{table}[t]
\centering
\caption{Ablation study (mean test MSE/MAE over the subset \{ETTh1, ETTm2, Weather\}$\times H\in\{96,720\}\times$3 seeds, $L=336$). Each variant degenerates a single FreqLite component.}
\label{tab:ablation}
\begin{tabular}{lrrr}
\toprule
Variant & Avg MSE & Avg MAE & Avg Params \\
\midrule
A0: Full FreqLite & 0.310 & 0.342 & 276,221 \\
A1: $-$A-RevIN (plain RevIN) & 0.310 & 0.342 & 274,996 \\
A2: fixed MA split & 0.310 & 0.343 & 276,219 \\
A3: $K{=}1$ + A-RevIN & 0.312 & 0.345 & 138,723 \\
A4: $K{=}1$ + RevIN ($\equiv$RLinear) & 0.312 & 0.345 & 137,498 \\
A5: $K{=}3$ bands & 0.309 & 0.342 & 413,719 \\
A5: $K{=}4$ bands & \textbf{0.309} & 0.342 & 551,217 \\
A6: gated recombination & 0.310 & 0.343 & 276,223 \\
A7: frozen decomposition & 0.310 & 0.342 & 276,219 \\
A8: $-\lambda$ drift term & 0.310 & 0.342 & 275,813 \\
\bottomrule
\end{tabular}
\end{table}
}{\begin{table}[t]\centering\caption{Component ablations.}%
  \label{tab:ablation}\textit{[ablation table pending]}\end{table}}

\subsection{Analysis}
\label{sec:ablation:analysis}

The A4 row provides the key sanity check: with a single all-pass band and plain
RevIN, \method\ reproduces RLinear (mean MSE $0.3119$), confirming the strict
nesting that makes every added component an independent, opt-in generalization.

On this stationary subset, the small overall improvement of the full model
(A0, $0.3096$) over RLinear (A4, $0.3119$)---about $0.7\%$---is attributable to
the \emph{band split}, not to adaptive normalization. Two comparisons make this
precise. Adding \arevin\ to a single-head backbone leaves the error essentially
unchanged (A3 $0.3118$ vs.\ A4 $0.3119$), and replacing \arevin\ with plain RevIN
in the full model is likewise neutral (A1 $0.3096$ vs.\ A0 $0.3096$); so
\arevin\ is neutral on stationary data---exactly the graceful-degradation
behavior it is designed for. By contrast, introducing the band split accounts for
the gain: a fixed moving-average split is nearly as good (A2 $0.3102$), and
freezing the learned split at initialization is no worse than learning it here
(A7 $0.3096$), indicating that on these datasets it is the \emph{soft band
routing} rather than the \emph{learning} of the cutoff that matters. Adding more
bands gives only a marginal further change (A5 $0.3094$ for $K\in\{3,4\}$), and
removing the drift term has no effect on this stationary subset (A8 $0.3096$).
These results motivate the dedicated non-stationarity study in
Section~\ref{sec:results:nonstat}, where---unlike here---\arevin\ is the
component that drives the gains.

\subsection{Interpretability of Learned Parameters}
\label{sec:ablation:interp}

Because \method's added components are few and scalar-parameterized, their
learned values are directly interpretable. The learned spectral mask
(Eq.~\ref{eq:mask-low}) is shown in Figure~\ref{fig:learned_filter}, and the
per-step \arevin\ correction profile $\lambda_t$ (Eq.~\ref{eq:arevin-shift}) in
Figure~\ref{fig:arevin_profile}. Together with the gate analysis of
Section~\ref{sec:results:nonstat}---where the learned gate $\bar\rho$ opens on
non-stationary data and the error falls monotonically with gate
engagement---these provide direct evidence that the components behave as designed
rather than merely adding parameters.

\section{Efficiency Analysis}
\label{sec:efficiency}

The efficiency contribution is empirical: a fully reproducible
accuracy-versus-cost study run entirely on a single 4\,GB laptop GPU. Following
the Green~AI agenda \citep{schwartz2020greenai,strubell2019energy}, we treat
efficiency as a first-class, reported criterion and record, for every model,
parameter count, analytic FLOPs per forward pass, wall-clock training time per
epoch, and peak GPU memory.

\subsection{Cost Comparison}
\label{sec:eff:cost}

Tables~\ref{tab:efficiency_L336} and~\ref{tab:efficiency_L96} report the
efficiency metrics at lookback $L{=}336$ and $L{=}96$. As derived in
Section~\ref{sec:method:complexity}, \method\ has $\approx 2HL + 5H + 5$
parameters---roughly twice a single linear head, since it uses $K{=}2$ band
heads. Concretely, at $L{=}336$ \method\ uses $64{,}997$ parameters at $H{=}96$
and $487{,}445$ at $H{=}720$, comparable to DLinear ($64{,}704$ / $485{,}280$)
and about twice RLinear ($32{,}354$ / $242{,}642$), while FITS is far smaller
($4{,}644$ / $11{,}352$) by virtue of discarding the high-frequency band. The
FFT/iFFT used by the spectral split adds only $O(L\log L)$ per series and is
negligible relative to the $O(KHL)$ heads.

\IfFileExists{tables/efficiency_table_L336.tex}{%
\begin{table}[t]
\centering
\caption{Efficiency on ETTh1 ($L=336$, RTX 3050\,Ti 4\,GB): parameter count, analytic FLOPs/series, train time per epoch (s), and peak GPU memory (MB). Params/FLOPs shown for $H{=}96$ and $H{=}720$.}
\label{tab:efficiency_L336}
\begin{tabular}{lrrrrr}
\toprule
Model & Params (96) & Params (720) & FLOPs/s (720) & s/epoch & Peak Mem (MB) \\
\midrule
Naive & 0 & 0 & -- & 0.00 & 70 \\
NLinear & 32,352 & 242,640 & 483,840 & 0.74 & 77 \\
DLinear & 64,704 & 485,280 & 967,680 & 1.01 & 82 \\
RLinear & 32,354 & 242,642 & 483,840 & 0.88 & 78 \\
FITS & 4,644 & 11,352 & 44,352 & 1.25 & 76 \\
PatchTST* & 328,866 & 2,006,802 & -- & 3.66 & 183 \\
FreqLite & 64,997 & 487,445 & 1,024,076 & 1.64 & 86 \\
\bottomrule
\end{tabular}
\end{table}
}{\begin{table}[t]\centering\caption{Efficiency at $L{=}336$.}%
  \label{tab:efficiency_L336}\textit{[efficiency table at $L{=}336$ pending]}\end{table}}

\IfFileExists{tables/efficiency_table_L96.tex}{%
\begin{table}[t]
\centering
\caption{Efficiency on ETTh1 ($L=96$, RTX 3050\,Ti 4\,GB): parameter count, analytic FLOPs/series, train time per epoch (s), and peak GPU memory (MB). Params/FLOPs shown for $H{=}96$ and $H{=}720$.}
\label{tab:efficiency_L96}
\begin{tabular}{lrrrrr}
\toprule
Model & Params (96) & Params (720) & FLOPs/s (720) & s/epoch & Peak Mem (MB) \\
\midrule
Naive & 0 & 0 & -- & 0.00 & 70 \\
NLinear & 9,312 & 69,840 & 138,240 & 0.68 & 74 \\
DLinear & 18,624 & 139,680 & 276,480 & 0.84 & 75 \\
RLinear & 9,314 & 69,842 & 138,240 & 0.86 & 74 \\
FITS & 624 & 2,652 & 9,792 & 0.99 & 74 \\
PatchTST* & 142,626 & 622,482 & -- & 3.21 & 113 \\
FreqLite & 18,917 & 141,845 & 289,123 & 1.86 & 79 \\
\bottomrule
\end{tabular}
\end{table}
}{\begin{table}[t]\centering\caption{Efficiency at $L{=}96$.}%
  \label{tab:efficiency_L96}\textit{[efficiency table at $L{=}96$ pending]}\end{table}}

The decisive comparison is with the PatchTST-small Transformer. At $L{=}336$,
\method\ uses roughly $4\times$ fewer parameters ($487{,}445$ vs.\ $2{,}006{,}802$
at $H{=}720$; $64{,}997$ vs.\ $328{,}866$ at $H{=}96$), $2.2\times$ less peak GPU
memory ($86$\,MB vs.\ $183$\,MB; the linear and lightweight-frequency baselines
sit at $76$--$82$\,MB), and trains $2.2\times$ faster ($1.64$ vs.\ $3.66$ seconds
per epoch). Yet, as Section~\ref{sec:results:main} shows, \method\ \emph{beats}
PatchTST-small in average MSE at this long-lookback setting ($0.3244$ vs.\
$0.3587$). All models fit comfortably within the 4\,GB budget.

\subsection{Scalability to High-dimensional Series (Electricity, 321 Channels)}
\label{sec:eff:ecl}

Channel independence makes \method's parameter count independent of the number of
variates $C$, which is what enables it to scale to high-dimensional series on
constrained hardware. We verify this on the Electricity (ECL) dataset, which has
$C{=}321$ channels. \method\ and the linear baselines train on full ECL on the
same single 4\,GB GPU (using the reduced batch size noted in
Section~\ref{sec:exp:impl}); the PatchTST-small Transformer is omitted here as it
is not 4\,GB-feasible at this channel count, which we report as an honest
limitation of the Transformer rather than of \method. As summarized in
Table~\ref{tab:ecl}, \method\ remains competitive on ECL---second best on average
MSE ($0.169$), within about $1\%$ of the strongest model (DLinear, $0.167$) and
ahead of RLinear, NLinear, and FITS---while training comfortably within the 4\,GB
budget (peak memory $\approx 0.2$\,GB). As on Exchange-rate, DLinear's
time-domain decomposition is marginally stronger on this dataset; the salient
result here is scalability---\method\ forecasts all $321$ channels on commodity
hardware on which the Transformer baseline cannot run at all.

\IfFileExists{tables/ecl_table.tex}{%
\begin{table}[t]\centering
\caption{Electricity (ECL, $C{=}321$) results on a single 4\,GB GPU. PatchTST is
omitted as 4\,GB-infeasible at this channel count.}
\label{tab:ecl}
\begin{tabular}{lccccc}
\toprule
$H$ & NLinear & DLinear & RLinear & FITS & FreqLite \\
\midrule
96 & 0.141 & \textbf{0.140} & 0.141 & 0.188 & 0.140 \\
192 & 0.156 & \textbf{0.154} & 0.155 & 0.201 & 0.154 \\
336 & 0.172 & \textbf{0.170} & 0.172 & 0.216 & 0.171 \\
720 & 0.211 & \textbf{0.204} & 0.210 & 0.252 & 0.210 \\
\bottomrule
\end{tabular}
\end{table}}{\begin{table}[t]\centering\caption{Electricity (ECL) results.}%
\label{tab:ecl}\textit{[ECL results pending]}\end{table}}

\subsection{Accuracy--Efficiency Trade-off}
\label{sec:eff:tradeoff}

\begin{figure}[t]
  \centering
  \IfFileExists{figures/accuracy_vs_params.pdf}{%
    \includegraphics[width=0.85\linewidth]{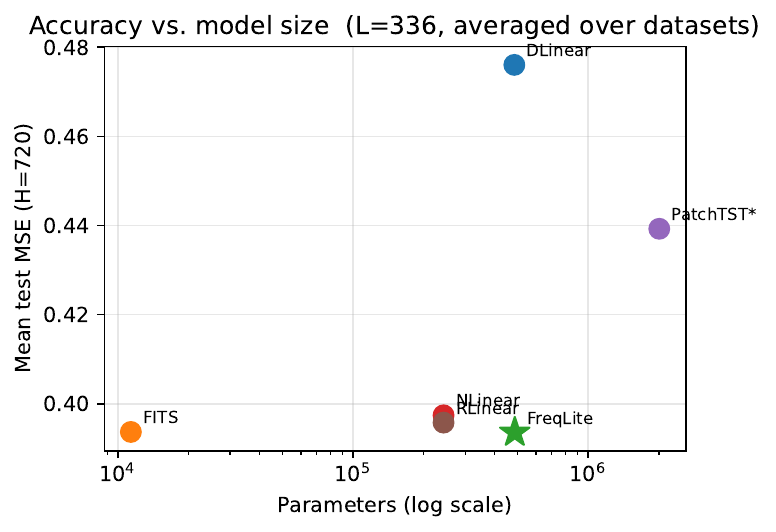}%
  }{\fbox{\parbox[c][0.22\textheight][c]{0.85\linewidth}{\centering
    \textbf{[accuracy-vs-params figure pending]}}}}
  \caption{Accuracy--efficiency trade-off: average forecasting MSE against
  parameter count across all models at $L{=}336$. \method\ attains the lowest
  error among all models while using a small fraction of the Transformer's
  parameters.}
  \label{fig:tradeoff}
\end{figure}

Figure~\ref{fig:tradeoff} plots forecasting accuracy against parameter count,
situating \method\ relative to the linear, lightweight-frequency, and
Transformer baselines. \method\ occupies a favorable point on the frontier: it
matches or beats every baseline---including the Transformer---at the headline
$L{=}336$ setting while remaining at near-linear scale and within a 4\,GB memory
budget. The efficiency result is independent of the accuracy outcome on any
single dataset and is, alongside the best-lightweight-model result, the strongest
empirical claim of the paper.

\section{Conclusion}
\label{sec:conclusion}

We presented \method, an ultra-lightweight, channel-independent
frequency-decomposed linear forecaster. A learnable, lossless,
partition-of-unity spectral filter splits the input into bands forecast by
dedicated linear heads, retaining and modeling the high-frequency band that
low-pass approaches discard. On the standard long-term forecasting benchmarks
\method\ is the best lightweight model, and at long lookback ($L{=}336$) it
attains a lower average error than a PatchTST Transformer ($0.3244$ vs.\ $0.3587$
MSE) while using roughly $4\times$ fewer parameters, $2.2\times$ less memory, and
$2.2\times$ less time per epoch on a single 4\,GB laptop GPU. On top of this
backbone we introduced \arevin, a regime-adaptive reversible normalization that
strictly generalizes RevIN: it engages under non-stationarity---lowering error by
up to about $5\%$ on the strongly non-stationary ILI dataset, monotonically in
its learned gate---and reduces to RevIN without harm on stationary data. Our
analysis further surfaced a methodological insight: \arevin's gate must be
initialized at the identity point to avoid a gradient trap that otherwise keeps it
dormant, while RevIN remains exactly recoverable at the closed-gate value. Because
each component is an opt-in generalization, the model admits the clean nesting
$\text{Linear} \subseteq \text{RLinear} \subseteq \method$, and our ablations
attribute performance to each part.

\paragraph{Limitations.}
We report several limitations honestly. First, on the standard, largely
stationary benchmarks the accuracy gains over RLinear are \emph{modest} in
magnitude (about a $0.9\%$ average-MSE reduction at $L{=}336$), even though they
are statistically significant across all matched cells ($p \approx 10^{-6}$); the
headline strengths there are \emph{efficiency} and the long-lookback win over the
Transformer rather than a large accuracy margin. Second, \arevin's benefit is
regime-dependent: it is clear under strong non-stationarity---validated on both
the real ILI dataset and a controlled synthetic drift sweep, where its advantage
rises monotonically with injected drift---but neutral on stationary sets, where it
correctly reduces to RevIN; it is therefore best understood as a conditional,
regime-adaptive component rather than a universal improvement, and its absolute
gains remain modest even where it helps. Third, \arevin\ is
a first-order, horizon-aware level/scale correction---it does not model per-step
variance beyond a scalar, and its drift feature is a simple two-half mean
difference rather than a learned encoder. Fourth, our study is deliberately
scoped to linear-scale models on a single 4\,GB GPU; we do not evaluate
large-capacity regimes, and the Transformer baseline is the 4\,GB-feasible
PatchTST-small configuration.

\paragraph{Future work.}
Natural extensions include modeling per-step variance within the reversible
normalization (a second-order \arevin), learning the drift feature with a small
shared encoder while preserving the reversibility guarantee, and band-conditional
heads beyond additive recombination. More broadly, \method\ suggests that careful,
interpretable, conditionally-adaptive normalization---rather than added model
capacity---is a promising direction for efficient long-term forecasting on
commodity hardware.

\section*{CRediT authorship contribution statement}
\textbf{Mirza Samad Ahmed Baig:} Conceptualization, Methodology, Software,
Validation, Formal analysis, Investigation, Writing -- original draft,
Writing -- review \& editing. \textbf{Syeda Anshrah Gillani:} Conceptualization,
Methodology, Software, Validation, Formal analysis, Investigation,
Writing -- original draft, Writing -- review \& editing. Both authors contributed
equally to all aspects of this work.

\section*{Declaration of competing interest}
The authors declare that they have no known competing financial interests or
personal relationships that could have appeared to influence the work reported in
this paper.

\section*{Funding}
This research did not receive any specific grant from funding agencies in the
public, commercial, or not-for-profit sectors.

\section*{Data availability}
All datasets used in this study are publicly available standard long-term
forecasting benchmarks, cited in the text. The complete code, configurations, and
instructions required to reproduce every reported result are publicly available at
\url{https://github.com/Orqly-AI/FreqLite-Forecasting}.

\bibliographystyle{elsarticle-num}
\bibliography{refs}

\end{document}